\documentclass[10pt,twocolumn,letterpaper]{article}

\usepackage{iccv}
\usepackage{times}
\usepackage{amsmath}
\usepackage{cases}
\usepackage{epsfig}
\usepackage{graphicx}
\usepackage{amssymb}
\usepackage{minitoc}
\usepackage{booktabs}
\usepackage{tabularray}
\usepackage{multirow}
\usepackage{mathtools}
\usepackage{pifont}
\usepackage{makecell}
\usepackage{subcaption}
\usepackage{algorithm}
\usepackage{algpseudocode}

\usepackage{amsmath,amsfonts,bm, bbm}

\def\eqref#1{equation~\ref{#1}}

\def\1{\bm{1}}

\DeclareMathAlphabet{\mathsfit}{\encodingdefault}{\sfdefault}{m}{sl}
\SetMathAlphabet{\mathsfit}{bold}{\encodingdefault}{\sfdefault}{bx}{n}

\def\gA{{\mathcal{A}}}

\def\gD{{\mathcal{D}}}

\def\gI{{\mathcal{I}}}

\def\gM{{\mathcal{M}}}

\def\gO{{\mathcal{O}}}
\def\gP{{\mathcal{P}}}

\def\gS{{\mathcal{S}}}

\newcommand{\vect}[1]{\ensuremath{\mathbf{#1}}}

\newcommand{\norm}[1]{\left\lVert#1\right\rVert}
\DeclarePairedDelimiterX{\infdivx}[2]{(}{)}{%
  #1\;\delimsize\|\;#2%
}

\newcommand{\xT}{\vect{x}_T}
\newcommand{\xt}{\vect{x}_t}
\newcommand{\xzero}{\vect{x}_0}
\newcommand{\xtone}{\vect{x}_{t-1}}
\newcommand{\zadv}{\vect{z}_\text{adv}}
\newcommand{\z}{\vect{z}}

\newcommand{\cmark}{\text{\ding{51}}}
\newcommand{\xmark}{\text{\ding{55}}}
\newcommand{\asr}[1]{{\color{magenta}#1}}
\newcommand{\fid}[1]{{\color{cyan}#1}}
\algnewcommand{\LineComment}[1]{\State \(\triangleright\) #1}

\usepackage[pagebackref=true,breaklinks=true,letterpaper=true,colorlinks,bookmarks=false]{hyperref}

\usepackage[capitalize]{cleveref}
\crefname{section}{Sec.}{Secs.}
\Crefname{section}{Section}{Sections}
\Crefname{table}{Table}{Tables}
\crefname{table}{Tab.}{Tabs.}

\iccvfinalcopy 

\def\iccvPaperID{3220} 

\ificcvfinal\fi

\begin{document}

\title{DiffProtect: Generate Adversarial Examples with Diffusion Models for Facial Privacy Protection}

\author{Jiang Liu$^{1}$, Chun Pong Lau$^{2}$, Zhongliang Guo$^{3}$, Yuxiang Guo$^{1}$, Zhaoyang Wang$^{1}$, Rama Chellappa$^{1}$\\
{\small $^{1}$Johns Hopkins University, $^{2}$City University of Hong Kong, $^{3}$University of St Andrews}}
\maketitle

\maketitle
\ificcvfinal\thispagestyle{empty}\fi

\begin{abstract}
The increasingly pervasive facial recognition (FR) systems raise serious concerns about personal privacy, especially for billions of users who have publicly shared their photos on social media. Several attempts have been made to protect individuals from being identified by unauthorized FR systems utilizing adversarial attacks to generate encrypted face images. However, existing methods suffer from poor visual quality or low attack success rates, which limit their utility. Recently, diffusion models have achieved tremendous success in image generation. In this work, we ask: can diffusion models be used to generate adversarial examples to improve both visual quality and attack performance? We propose DiffProtect, which utilizes a diffusion autoencoder to generate semantically meaningful perturbations on FR systems. Extensive experiments demonstrate that DiffProtect produces more natural-looking encrypted images than state-of-the-art methods while achieving significantly higher attack success rates, \eg, 24.5\% and 25.1\% absolute improvements on the CelebA-HQ and FFHQ datasets. 
\vspace{-5mm}
\end{abstract}

\setlength{\abovedisplayskip}{1pt}
\setlength{\belowdisplayskip}{1pt}
\setlength{\abovecaptionskip}{0pt}
\setlength{\belowcaptionskip}{0pt}

\section{Introduction}
\looseness -1 The rise of deep neural networks has enabled the tremendous success of facial recognition (FR) systems~\cite{deng2019arcface, schroff2015facenet, mobileface}. However, the widely deployed FR systems also pose a huge threat to personal privacy as billions of users have publicly shared their photos on social media. Through large-scale social media photo analysis, FR systems can be used for detecting user relationships~\cite{10.1515/popets-2015-0004},  stalking victims~\cite{shwayder2020clearview}, stealing identities~\cite{lively}, and performing massive government surveillance~\cite{satariano2019police, hill2020secretive, mozur2019surveillance}. It is urgent to develop facial privacy protection techniques to protect individuals from unauthorized FR systems.

\begin{figure}[t]
  \centering
\includegraphics[width=0.5\textwidth]{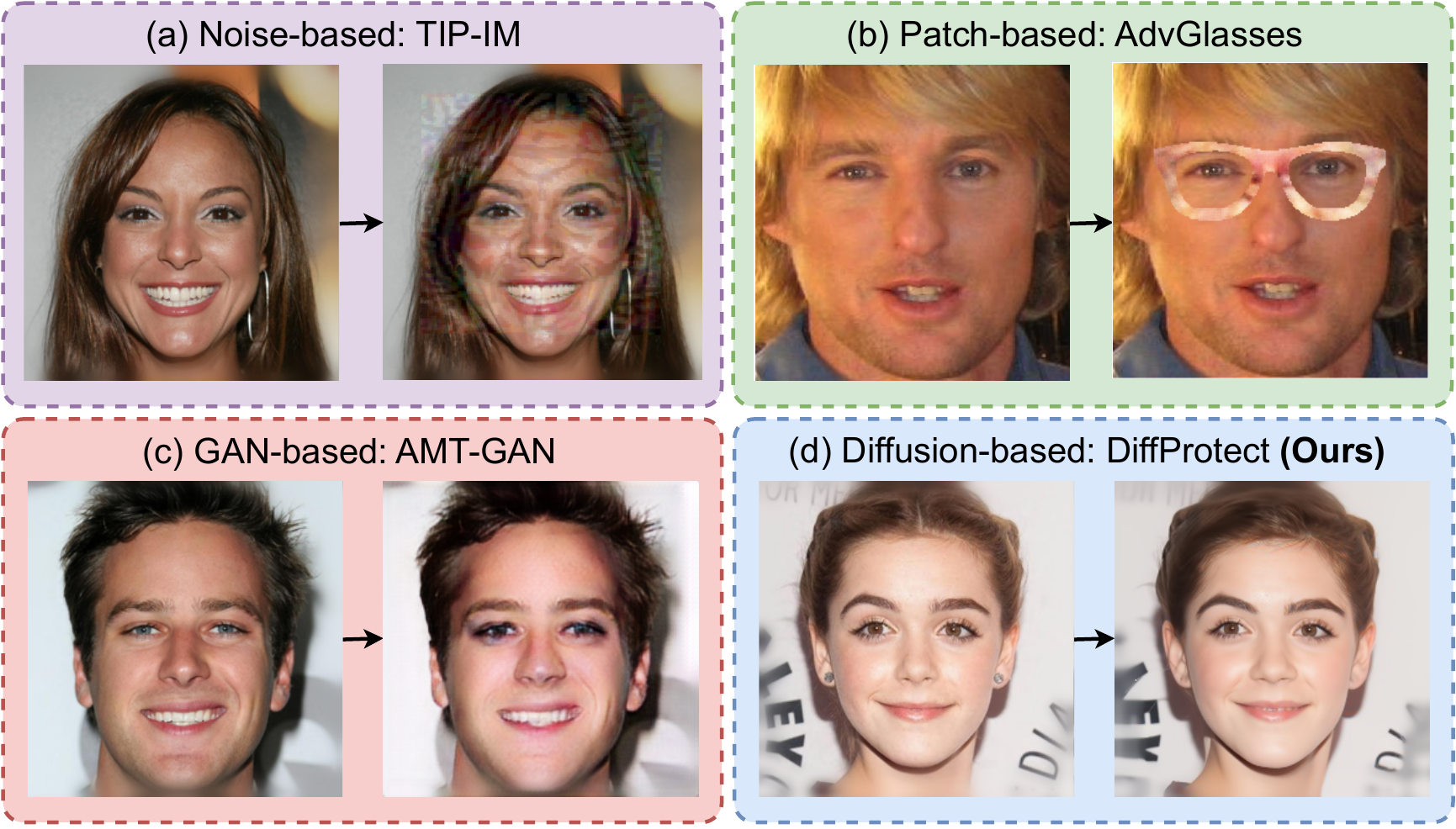}
  \caption{Illustration of different face encryption methods~\cite{yang2021towards, hu2022protecting, sharif2019general}. The proposed DiffProtect produces natural and imperceptible changes to the input images while achieving competitive attack success rates. }
  \label{fig:teaser}
\end{figure}

Recently, several works~\cite{cherepanova2021lowkey, yang2021towards, hu2022protecting} proposed to use adversarial attacks to generate encrypted face images and protect users from being identified by FR systems. However, existing attacks on FR systems~{\cite{cherepanova2021lowkey, hu2022protecting, komkov2021advhat, sharif2019general, yin2021adv, yang2021towards, zhong2020towards}} often suffer from poor visual quality, especially noise-based~\cite{cherepanova2021lowkey, yang2021towards} (\cref{fig:teaser}a) and patch-based methods~\cite{komkov2021advhat, sharif2019general} (\cref{fig:teaser}b), which add unnatural and conspicuous changes to the source images. From a user-centered perspective, these approaches are not desirable in practice, as everyone wants to post their best-looking photos on social media -- not unattractive photos, even though they might be encrypted for protecting personal privacy. 

An ideal facial identity encryption algorithm should only create \textit{natural} or \textit{imperceptible} changes to the source images. To achieve this goal, several works~\cite{yin2021adv, hu2022protecting, qiu2019semanticadv, choi2018stargan} attempted to generate natural-looking adversarial examples using generative adversarial networks (GANs)~\cite{gan, karras2019style} by transferring make-up styles~\cite{yin2021adv, hu2022protecting} or editing facial attributes~\cite{qiu2019semanticadv, jia2022advattribute}. Although these works have demonstrated promising results, the attack success rates (ASRs) of GAN-based methods can be lower than noise-based and patch-based methods since the output images are constrained on the learned manifold of a GAN. Moreover, most of the GAN-based facial encryption methods~\cite{hu2022protecting, yin2021adv} require training the model with a fixed victim identity. In other words, these models are target specific and we need to retrain the model when the target identity is changed. In addition, the visual quality of images generated by GAN-based methods is still dissatisfying (see~\cref{fig:teaser}c). 

Recently, diffusion models~\cite{ho2020denoising, song2020score, song2020denoising, diffae, Karras2022edm} have emerged as state-of-the-art generative models that are capable of synthesizing realistic and high-resolution images. It has been shown that diffusion models perform better than GANs on many image generation tasks~\cite{guided_diffusion, stablediffusion, diffae} and maintain better coverage of the image distribution~\cite{guided_diffusion}. With the rise of diffusion models, a natural question arrives: \textit{can we utilize diffusion models as better generative models to generate adversarial examples with both high attack success rates and high visual quality?} It has been shown that diffusion models can improve adversarial robustness~\cite{gowal2021generated, rebuffi2021fixing, wang2023better, nie2022DiffPure, carlini2023certified}. However, whether the strong generative capacity of diffusion models can be used to generate adversarial attacks has not been explored.

To this end, we propose DiffProtect, which utilizes a pretrained diffusion autoencoder~\cite{diffae} to generate adversarial images for facial privacy protection. The overall pipeline of DiffProtect is shown in~\cref{fig:pipeline}. We first encode an input face image $\vect{I}$ into a high-level semantic code $\vect{z}$ and a low-level noise code $\xT$. We then iteratively optimize an adversarial semantic code $\vect{z}_{\text{adv}}$ such that the resulting protected image $\vect{I}_p$ generated by the conditional DDIM decoding process~\cite{diffae, song2020denoising} can fool the face recognition model. In this way, we can create semantically meaningful perturbation to the input image and utilize a diffusion model to generate high-quality adversarial images (see~\cref{fig:teaser}d). We further introduce a face semantics regularization module to encourage that $\vect{I}_{\text{p}}$ and $\vect{I}$ share similar face semantics to better preserve visual identity. Our extensive experiments on the CelebA-HQ~\cite{karras2017progressive} and FFHQ~\cite{karras2019style} datasets show that DiffProtect produces protected face images of high visual quality, while achieving significantly higher ASRs than the previous best methods, \eg, + 24.5\% absolute ASR on CelebA-HQ with IRSE50~\cite{hu2018squeeze} as the victim model.  

One drawback of diffusion models is the slow generation process, which requires many evaluation steps to generate one image. This can be problematic since we need to iteratively optimize the semantic latent code. To address this issue, we propose an attack acceleration strategy that computes an approximated version of the reconstructed image at each attack iteration by running only one generative step, which drastically reduces the attack time while maintaining competitive attack performance.

Our main contributions are summarized as follows:
\begin{itemize}
    \item We propose a novel diffusion model based attack method, termed DiffProtect, for facial privacy protection, which crafts natural and inconspicuous adversarial examples on face recognition systems.  
    \item We further propose a face semantics regularization module to better preserve visual identity and a simple yet effective attack acceleration strategy to improve attack efficiency.
    \item Our extensive experiments on the CelebA-HQ and FFHQ datasets demonstrate that DiffProtect produces more natural-looking encrypted images than state-of-the-art methods while achieving competitive attack performance. 
\end{itemize}
\section{Related Work}

\paragraph{Adversarial Attacks with Generative Models} \looseness -1 While most of the existing adversarial attacks focus on optimizing additive noises in the pixel space~\cite{goodfellow2015explaining, madry2017towards, carlini2017towards, dong2018boosting}, several works~\cite{wong2021learning, xiao2018generating, jalal2017robust, qiu2019semanticadv, yin2021adv, hu2022protecting} proposed to generate adversarial attacks with generative models, which produce perceptually realistic adversarial examples. Wong \etal ~\cite{wong2021learning} trained a conditional variational autoencoder (VAE)~\cite{Sohn2015LearningSO, vae} to generate a variety of perturbations and utilized the learned perturbation sets to improve model robustness and generalization. Xiao \etal~\cite{xiao2018generating} trained a conditional GAN to directly produce adversarial examples.~\cite{jalal2017robust, lin2020dual, Stutz_2019_CVPR} generate on-manifold adversarial examples by optimizing latent codes in the latent space of GANs. Qiu~\etal~\cite{qiu2019semanticadv} generated semantically realistic adversarial examples by attribute-conditioned image editing using a GAN. To the best of our knowledge, this is the first work that utilizes diffusion models for generating adversarial attacks. 

\paragraph{Diffusion Models and Adversarial Robustness} Diffusion models~\cite{sohl2015deep, ho2020denoising, nichol2021improved} are a family of generative models that model the target distribution by learning a reverse generative denoising process. Recently, diffusion models have become state-of-the-art methods that can synthesize more realistic and high-resolution images with more stable training process~\cite{ho2020denoising, song2020score, song2020denoising, diffae, Karras2022edm}. 
In the field of adversarial machine learning, it has been shown that diffusion models can help to improve adversarial robustness and achieve state-of-the-art results~\cite{gowal2021generated, rebuffi2021fixing, wang2023better, nie2022DiffPure, carlini2023certified}.~\cite{gowal2021generated, rebuffi2021fixing, wang2023better} used diffusion models to generate synthetic data for adversarial training. ~\cite{nie2022DiffPure, carlini2023certified} exploited pretrained diffusion models as purification modules that remove the adversarial noise in the input images for empirical~\cite{nie2022DiffPure} and certified~\cite{carlini2023certified} defenses. However, it remains unknown whether the strong generative capacity of diffusion models can be used to generate adversarial attacks and in this work we make the first attempt in this direction.

\paragraph{Adversarial Attacks on Face Recognition}
Many studies have been proposed to attack FR systems, including both poisoning~\cite{shan2020fawkes} and evasion~\cite{cherepanova2021lowkey, hu2022protecting, komkov2021advhat, sharif2019general, yin2021adv, yang2021towards} attacks. Poisoning attacks require injecting poisoned face images into the training sets of FR systems, which is unlikely to achieve for individual users. Evasion attacks, especially transferable black-box attacks, are more practical for protecting facial image privacy, as they only require perturbing the source images to fool the FR systems at test time. Existing attacks on FR systems~\cite{cherepanova2021lowkey, hu2022protecting, komkov2021advhat, sharif2019general, yin2021adv, yang2021towards} suffer from poor visual quality or low attack success rates, which limit their usability in the real world. Noise-based methods such as Lowkey~\cite{cherepanova2021lowkey} and TIP-IM~\cite{yang2021towards} create unexplainable noise patterns on face images. Patch-based methods such as Adv-Hat~\cite{komkov2021advhat} and Adv-Glasses~\cite{sharif2019general} add unnatural and conspicuous patches to the source images. GAN-based methods generate more natural-looking adversarial examples~\cite{yin2021adv, hu2022protecting, qiu2019semanticadv, jia2022advattribute}, but typically have lower attack success rates. In addition,~\cite{yin2021adv, hu2022protecting} drastically change the makeup styles of source images and might be biased towards female users~\cite{hu2022protecting}. In contrast, the proposed DiffProtect creates natural and imperceptible changes to the source images and achieves competitive attack performance.

\section{Background: Diffusion Models}
\begin{figure*}[t]
  \centering
\includegraphics[width=0.95\textwidth]{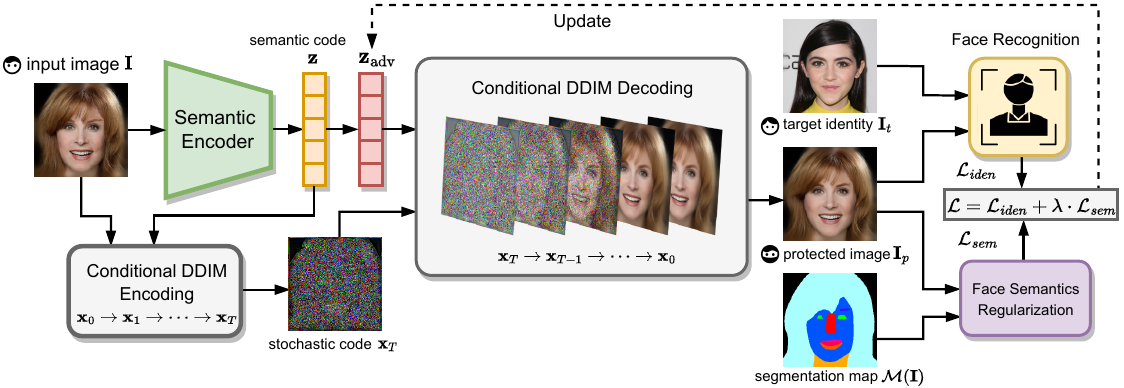}
  \caption{Overview of DiffProtect. DiffProtect utilizes a diffusion autoencoder~\cite{diffae}, which encodes the input image as a semantic code $\z$ and a stochastic code $\xT$. We iteratively optimize an adversarial semantic code $\vect{z}_{\text{adv}}$ such that the resulting protected image $\vect{I}^p$ generated by the conditional DDIM decoding process can fool the face recognition model. }
  \label{fig:pipeline}
\end{figure*}
Diffusion models~\cite{sohl2015deep, ho2020denoising, nichol2021improved} are a family of generative models that model the target distribution by learning a reverse denoising generative process. A diffusion model consists of two processes: (1) a forward diffusion process that converts the input image $\xzero$ to noise map $\xT$ by gradually adding noise in $T$ forward steps; (2) a reverse denoising generative process that aims to recover $\xzero$ from $\xT$ by gradually denoising in $T$ reverse steps. 

A Gaussian diffusion process gradually adds Gaussian noise to the data in the forward diffusion process:
\begin{equation}
    q(\xt|\xtone) = \mathcal{N}(\sqrt{1 - \beta_t}\xtone, \beta_t \mathbf{I}),
\end{equation} 
where $\beta_t\ (t=1,..., T)$ are hyperparameters controlling the noise level at each diffusion step $t$. In the Gaussian diffusion process, the noisy image $\xt$ also follows a Gaussian distribution given $\xzero$:
\begin{equation}
    \label{eq:margin}
    q(\xt|\xzero) = \mathcal{N}(\sqrt{\alpha_t}\xzero, (1 - \alpha_t) \vect{I}),
\end{equation} 
where $\alpha_t = \prod_{s=1}^t (1-\beta_s)$. The goal of a diffusion model is to learn the reverse distribution $p(\xtone|\xt)$ so that we can generate data by sampling noise map $\xT$ from a prior distribution, \eg, $\mathcal{N}(\vect{0}, \mathbf{I})$. When the difference between $t-1$ and $t$ is infinitesimally small, \ie, $T=\infty$, $p(\xtone|\xt)$ can be modeled as~\cite{sohl2015deep, ho2020denoising}:

\begin{equation}
    p(\xtone|\xt) = \mathcal{N}(\mathbf{\mu_\theta}(\xt, t), \sigma_t^2 \mathbf{I}),
\end{equation}
where $\mu_\theta$ is a neural network that predicts the posterior mean given the noisy image $\xt$ and time step $t$. In practice, Ho \etal~\cite{ho2020denoising} proposed to train a U-Net~\cite{Unet} to learn a function $\epsilon_\theta(\xt, t)$ the noise that has been added to $\xzero$ by the following reweighted loss function: 
\begin{equation}
    \mathcal{L}_{\text{simple}}= \sum_{t=1}^T \mathbb{E}_{\xzero, \vect{\epsilon}_t}\left[\norm{\epsilon_t - \epsilon_\theta(\xt, t)}^2\right],
\end{equation}
where $\epsilon_t$ is the ground-truth noise added to $\xzero$ to produce $\xt$. And $\mu_\theta(\xt, t)$ can be obtained by:
\begin{equation}
    \mu_\theta(\xt, t) = \frac{1}{\sqrt{1-\beta_t}}\left(\xt - \frac{\beta_t}{\sqrt{1-\alpha_t}}\epsilon_\theta(\xt, t)\right).
\end{equation}

Song \etal~\cite{song2020denoising} proposed Denoising Diffusion Implicit Model (DDIM) that enjoys a deterministic generative process. DDIM has a non-Markovian forward process:
\begin{equation}
\resizebox{1\hsize}{!}{$
q(\xtone|\xt,\xzero) = \mathcal{N}\Big(\sqrt{\alpha_{t-1}} \xzero + \sqrt{1 - \alpha_{t-1}} \frac{\xt - \sqrt{\alpha_t} \xzero}{\sqrt{1 - \alpha_t}}, \vect{0} \Big)$}.
\label{eq:ddimq}
\end{equation}
In the reverse process, DDIM first predicts $\xzero$ given $\xt$: 
\begin{equation}
    f_{\theta}(\xt, t)=(\xt - \sqrt{1-\alpha_t }\cdot \epsilon_\theta(\xt, t))/\sqrt{\alpha_t}, 
\end{equation}
and the reverse process is given by replacing $\xzero$ with $f_{\theta}(\xt, t)$ in \cref{eq:ddimq}: 
\begin{equation}
\resizebox{1\hsize}{!}{$\xtone = \sqrt{\alpha_{t-1}} \left( \frac{\xt - \sqrt{1 - \alpha_t} \epsilon_\theta(\xt, t)}{\sqrt{\alpha_t}} \right) + \sqrt{1 - \alpha_{t-1}} \epsilon_\theta(\xt, t)$}.
\label{eq:gen}
\end{equation}
We can think of DDIM as an \textit{image encoder}, where we can run the generative process backward deterministically to obtain a noise map $\xT$ that serves as a latent variable representing $\xzero$, as well as an \textit{image decoder}, where we run the generative process (\cref{eq:gen}) to decoder $\xzero$ from $\xT$.
 
\section{DiffProtect}
\subsection{Problem Formulation}
In this section, we formulate the problem of adversarial attacks on face recognition systems. Suppose the face images $\vect{I}\in \gI := \mathbb{R}^{H \times W \times C}$ are drawn from an underlying distribution $\mathbb{P}_{\gI}$, where $H$, $W$ and $C$ are the height, width and the number of channels of the image respectively. Let $h(\vect{I}):\gI \rightarrow \mathbb{R}^d$ denote a face recognition model which maps an input face image in $\gI$ to a feature vector in $\mathbb{R}^d$. An \emph{accurate} FR system can map two images $\vect{I}_1$ and $\vect{I}_2$ of the same identity to features that are close in the feature space, and to features that are far away when they are of different identities, \ie, $\gD(h(\vect{I}_1), h(\vect{I}_2)) \leq \tau$ when $y_1 = y_2$ and $\gD(h(\vect{I}_1), h(\vect{I}_2)) > \tau$ when $y_1 \neq y_2$, where $\gD$ is a distance function, $\tau$ is a threshold and $y_1$, $y_2$ are the identity of $\vect{I}_1$ and $\vect{I}_2$ respectively. 

In the {untargeted} attack or ``dodging attack" setting, a successful attack fools the face recognition system to map an adversarial image $\vect{I}_\text{adv}$ with the same identity as $\vect{I}$ to a feature that is far away from $h(\vect{I})$, \ie,  $\gD(h(\vect{I}), h(\vect{I}_\text{adv})) > \tau$. In the {targeted} attack or ``impersonation attack" setting, a successful attack fools the face recognition system to map an adversarial image $\vect{I}_\text{adv}$ with the targeted identity to a feature that is close to $h(\vect{I})$, i.e. $\gD(h(\vect{I}), h(\vect{I}_\text{adv})) \leq \tau$. In this work, we focus on targeted attack setting following previous work~\cite{zhong2020towards, qiu2019semanticadv, hu2022protecting}.

\subsection{Detailed Construction}
The overall pipeline of DiffProtect is shown in~\cref{fig:pipeline}. DiffProtect consists of a semantic encoder and a conditional DDIM that serves as a stochastic encoder and an image decoder. An input face image is first encoded as a high-level semantic code $\z$ and a stochastic code $\xT$ that captures low-level variations. We aim to optimize an adversarial semantic code $\z_{\text{adv}}$ such that the resulting protected image $\vect{I}_{p}$ generated by the conditional DDIM decoding process~\cite{diffae, song2020denoising} can fool the face recognition model to protect facial privacy. 

\paragraph{Semantic Encoder} The semantic encoder learns to map an input face image $\vect{I}$ into a semantic latent code $\z= Enc(\vect{I})$ that captures high-level face semantics. Manipulating $\z$ results in semantic changes in the image.

\paragraph{Conditional DDIM} The conditional DDIM~\cite{diffae} is a DDIM model conditioned on the semantic code $\z$, where we train a noise prediction network $\epsilon_\theta(\xt, t, \z)$ with $\z$ as an additional input. During the decoding process, we obtain the reconstructed image $\vect{I}=\xzero = \text{DDIM}_{dec}(\xT, \z)$, by running the following deterministic generative process:
\begin{equation}
    p_\theta(\vect{x}_{t-1} | \xt, \z) = \begin{cases}
    \mathcal{N}(f_\theta(\vect{x}_1, 1, \z), \vect{0}) & \hspace{-0.27cm}\text{if } t = 1 \\
    q(\vect{x}_{t-1}|\vect{x}_t, {f}_\theta(\xt, t, \z)) & \hspace{-0.27cm}\text{otherwise},
  \end{cases}
    \label{eq:rev}
\end{equation}
where $f_{\theta}(\xt, t, \z)=(\xt - \sqrt{1-\alpha_t }\cdot \epsilon_\theta(\xt, t, \z))/\sqrt{\alpha_t}$, and $q(\cdot|\cdot, \cdot)$ is defined in~\cref{eq:ddimq}.

During the encoding process,  we obtain the stochastic code of the input image $ \xT = \text{DDIM}_{enc}(\vect{I}, \z)$ by running the deterministic generative process backward:
\begin{equation}
    \vect{x}_{t+1} = \sqrt{\alpha_{t+1}}{f}_\theta(\xt, t, \z) +\sqrt{1-\alpha_{t+1}}\vect{\epsilon}_\theta(\vect{x}_t, t, \z).
\end{equation}
$\xT$ is encouraged to encode only the information left
out by $\z$, \ie, the stochastic details.

\paragraph{Attack Formulation} To generate a protected image $\vect{I}_{p}$ that can effectively fool the face recognition systems and has a high visual quality to human eyes, we perturb the semantic code $\z$ of the input image $\vect{I}$ to obtain an adversarial semantic code $\zadv$ to create semantically meaningful perturbations. We then generate $\vect{I}_{p}$ by feeding $\zadv$ and $\xT$ to the DDIM decoding process: 
\begin{equation}
    \vect{I}_{p} = \text{DDIM}_{dec}(\xT, \zadv).
    \label{eq:Ip}
\end{equation}
In this way, we constrain $\vect{I}_{p}$ to lie on the manifold of real image distribution to ensure that $\vect{I}_{p}$ has high visual quality. 

Formally, for targeted privacy protection, we aim to solve the following optimization problem: 
\begin{equation}
    \min_{\zadv} \mathcal{L}_{iden}(\vect{I}_{p}) = \gD(h(\vect{I}_{p}), h(\vect{I}_{t})), \text{s.t.} \norm{\zadv - \z}_{\infty} < \gamma, 
    \label{eq:attack}
\end{equation} 
where $\vect{I}_{t}$ is the face image of the target identity, $\vect{I}_{p}$ is the protected face image generated as~\cref{eq:Ip}, $\gamma$ is the attack budget, $\gD$ is the cosine distance, $h$ is the face recognition model. In practice, the user may not have access to $h$. In such a black-box attack setting, we use surrogate models $g \in \gA_g$ to estimate the identity loss $\mathcal{L}_{iden}$:
\begin{equation}
    \mathcal{L}_{iden}(\vect{I}_{p}) = \sum_{g \in \gA_g} \gD(g(\vect{I}_{p}), g(\vect{I}_{t})).
\end{equation}

\paragraph{Face Semantics Regularization} Although the attack strength is controlled by a small attack budget $\gamma$ in~\cref{eq:attack} to preserve the overall facial identity to human eyes, in some cases the protected image can have different local facial features compared to the input image in order to match the characteristics of the target image, such as the face shape in Fig.~\ref{fig:face_parsing}, which may not be desired by some users. 

To regularize the face semantics of the protected image, we propose a face semantic consistency loss $\mathcal{L}_{sem}$ that computes the similarity between the semantic maps of the protected image  $\vect{I}_{p}$ and the input image $\vect{I}$:
\begin{equation}
    \mathcal{L}_{sem}(\vect{I}_{p}) = \text{CE}(\gM(\vect{I}_{p}), \gM(\vect{I})), 
\end{equation}
where $\gM$ is a face parsing network that outputs the semantic map of the input image, and CE is the cross-entropy loss. The optimization problem in~\cref{eq:attack} becomes:
\begin{equation}
    \min_{\zadv}\mathcal{L}(\vect{I}_{p}) = \mathcal{L}_{iden}(\vect{I}_{p}) + \lambda \cdot \mathcal{L}_{sem}(\vect{I}_{p}), \text{s.t.} \norm{\zadv - \z}_{\infty} < \gamma, 
    \label{eq:attack_total}
\end{equation} 
 where $\lambda$ is a hyper-parameter that controls the importance of the semantic consistency term. 
 
\vspace{-3mm}
\paragraph{Attack Generation} The DiffProtect algorithm is summarized in Alg.~\ref{alg:diffprotect}. We iteratively solve~\cref{eq:attack_total} with projected gradient descent~\cite{madry2017towards}:

\begin{numcases}{}
    \vect{I}_{p}^{(i)} = \text{DDIM}_{dec}(\xT, \zadv^{(i)})  \label{eq:Ipi} \\
    \zadv^{(i+1)}=\gP_{\gS}\left(\zadv^{(i)} - \eta \cdot \text{sign}\left(\nabla_{\zadv^{(i)}} \mathcal{L}(\vect{I}_{p}^{(i)})  \right)\right) \label{eq:pgd} 
\end{numcases}
where $\zadv^{(0)}=\z$, $\eta$ is the attack step size, $\gP_{\gS}$ is the projection onto feasible set $\gS=\{\zadv: \norm{\zadv - \z}_{\infty} < \gamma\}$. Note that the DDIM decoding process is fully differentiable (\cref{eq:rev}) so we can directly compute $\nabla_{\zadv} \mathcal{L}(\vect{I}_{p})$ to update $\zadv$.

\begin{algorithm}[tb]
\caption{DiffProtect}\label{alg:diffprotect}

\begin{algorithmic}[1]
\State {\bfseries Input:} $\vect{I}$, $N$, $\gamma$, $\eta$, $\lambda$, $T$, $Enc$, $\epsilon_\theta$, $h$ or $\gA_g$, $\gM$ %
\State {\bfseries Output:} protected image $\vect{I}_p$ 
\LineComment{Image Encoding}
\State $\z= Enc(\vect{I})$, $ \xT = \text{DDIM}_{enc}(\vect{I}, \z)$
\LineComment{Attack Generation}
\State $\zadv^{(0)} \gets \z$
\State {\bfseries for} $i=0$ {\bfseries to} $N-1$
\State $\ \ \ \ \vect{I}_{p}^{(i)} = \text{DDIM}_{dec}(\xT, \zadv^{(i)})$ 
\State $\ \ \ \ ${\bfseries or} $\vect{I}_{p}^{(i)} \approx f_{\theta}(\vect{x}_{t_0}, t_0, \zadv^{(i)})$ \Comment{DiffProtect-fast}
\State $\ \ \ \ \mathcal{L}(\vect{I}_{p}^{(i)}) = \mathcal{L}_{iden}(\vect{I}_{p}^{(i)}) + \lambda \cdot \mathcal{L}_{sem}(\vect{I}_{p}^{(i)})$
\State $\ \ \ \  \zadv^{(i+1)}=\gP_{\gS}\left(\zadv^{(i)} - \eta \cdot \text{sign}\left(\nabla_{\zadv^{(i)}} \mathcal{L}(\vect{I}_{p}^{(i)}) \right)\right) $
\State {\bfseries end for}
\State $\zadv \gets \zadv^{(N)}$, $\vect{I}_p = \text{DDIM}_{dec}(\xT, \zadv)$ 

\end{algorithmic}
\end{algorithm}

\paragraph{Attack Acceleration} In~\cref{eq:Ipi}, we need to run the full DDIM decoding process to obtain $\vect{I}_{p}^{(i)}$ at each attack iteration. The time complexity for generating a protected image is $\gO(T \times N)$, where $T$ is the number of generative steps in the DDIM decoding process, $N$ is the number of attack steps. To accelerate attack generation, we can instead compute an approximated version of $\vect{I}_{p}^{(i)}$ at each attack iteration by running only one generative step. Specifically, let $t_0$ be the time stamp from which we run the generative step to estimate $\vect{I}_{p}^{(i)}$. We can obtain the corresponding noise map $\vect{x}_{t_0}$ from the DDIM encoding process and estimate $\vect{I}_{p}^{(i)}$ by:
\begin{equation}
\small
    \vect{I}_{p}^{(i)} \approx f_{\theta}(\vect{x}_{t_0}, t_0, \zadv^{(i)}) = \frac{1}{\sqrt{\alpha_{t_0}}}(\vect{x}_{t_0} - \sqrt{1-\alpha_{t_0} }\cdot \epsilon_\theta(\vect{x}_{t_0}, t_0, \zadv^{(i)})).
    \label{eq:attack_fast}
\end{equation}
In this way, we can reduce the time complexity to $\gO(N)$. We denote this accelerated version of DiffProtect as DiffProtect-fast. Note that $\vect{x}_{t_0}$ is not updated during the attack. This simple strategy works surprisingly well as we will show in~\cref{sec:acceleration}. The effect of the choice of $t_0$ is shown in the supplementary materials.

\begin{table*}[t]
\centering
\setlength{\tabcolsep}{0.2mm}
\scalebox{0.94}{\setlength{\tabcolsep}{2.0mm}\begin{tabular}{l|cc|ccc|c|ccc|c}
\toprule[1pt]
\multirow{3}{*}{Methods}   & \multirow{3}{*}{\makecell{Target-\\ specific}}   & \multirow{3}{*}{\makecell{Natural-
\\looking}}     & \multicolumn{4}{c|}{CelebA-HQ}  & \multicolumn{4}{c}{FFHQ}  \\
\cline{4-11}
 & &  & \multicolumn{3}{c|}{ASR (\%) $\uparrow$} & \multirow{2}{*}{FID $\downarrow$} & \multicolumn{3}{c|}{ASR (\%) $\uparrow$} & \multirow{2}{*}{FID $\downarrow$} \\
 
\cline{4-6} \cline{8-10}
             &  &  & IRSE50    & IR152 & MobileFace &  & IRSE50 & IR152 & MobileFace &  \\
\hline
No attack  & $\xmark$  & $\cmark$ & 7.3      & 3.8   & 1.1   & 0    & 4.4    & 2.5   & 5.2    &  0 \\
PGD~\cite{madry2017towards}  & $\xmark$  & $\xmark$            & 32.6      & 19.0    & 35.6  & 39.6     & 20.1   & 14.4  & 18.8 & 38.1     \\
MIM~\cite{dong2018boosting}  & $\xmark$  & $\xmark$            & 37.4      & 31.0    & 35.5   & 69.9    & 24.5   & 23.2  & 21.4 & 62.1      \\
TIP-IM~\cite{yang2021towards}   & $\xmark$  & $\xmark$        & 47.2      & 35.3  & 45.3   & 71.6     & {31.0}     & 27.5  & 26.9   &   62.8  \\
AMT-GAN~\cite{hu2022protecting}  & $\cmark$  & $\cmark$        & {53.9}        & {41.9}  & {60.0}   &  {31.1}  &  {32.6}  &  30.5  & {29.9}  & {30.5} \\
\hline
DiffProtect-fast $(\lambda=0)$ & $\xmark$  & $\cmark$ & \underline{68.4}      & \underline{49.8}  & \underline{72.1}     &  \underline{26.7} & \underline{50.8}   & \underline{47.6}  & \underline{47.0} &   26.4\\
DiffProtect $(\lambda=0)$  & $\xmark$  & $\cmark$     & \textbf{78.4}      & \textbf{60.3}  & \textbf{77.9}  & {27.6}    & \textbf{57.7}   & \textbf{54.3}  & \textbf{52.9} & \underline{26.1} \\
DiffProtect $(\lambda=0.2)$ & $\xmark$  & $\cmark$ & 67.7      & 48.7  & {69.3}     & \textbf{24.4} & 46.2   & 45.4  & 44.3   & \textbf{23.5}    \\
\bottomrule[1pt]
\end{tabular}}
\caption{Comparison with state-of-the-art methods on CelebA-HQ and FFHQ datasets for targeted black-box attacks. The best performance is in \textbf{bold} and the second best is \underline{underlined}.}
\label{tab:asr_main}
\end{table*}
\begin{figure*}[t]
\captionsetup[subfigure]{labelformat=empty}
    \centering
    \small
    \setlength{\tabcolsep}{0.5mm}
\scalebox{0.98}{
\begin{tabular}[b]{cccccccc}
    \rotatebox{90}{\hskip 2.5em Input} & 
    \begin{subfigure}{0.14\textwidth}
        \includegraphics[width=\textwidth]{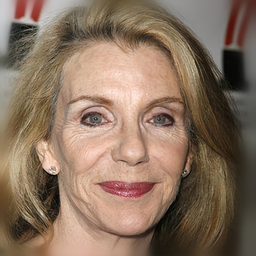}
    \end{subfigure} & 
    \begin{subfigure}{0.14\textwidth}
        \includegraphics[width=\textwidth]{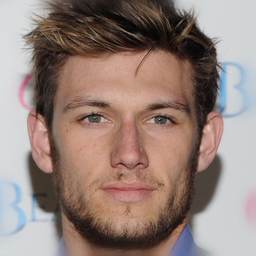}
    \end{subfigure} & 
    \begin{subfigure}{0.14\textwidth}
        \includegraphics[width=\textwidth]{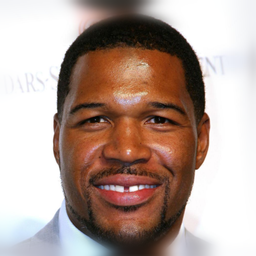}
    \end{subfigure} & 
    \begin{subfigure}{0.14\textwidth}
        \includegraphics[width=\textwidth]{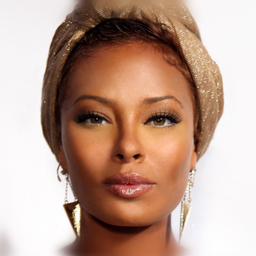}
    \end{subfigure} & 
    \begin{subfigure}{0.14\textwidth}
        \includegraphics[width=\textwidth]{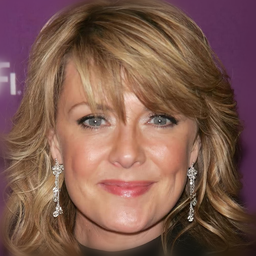}
    \end{subfigure} & 
    \begin{subfigure}{0.14\textwidth}
        \includegraphics[width=\textwidth]{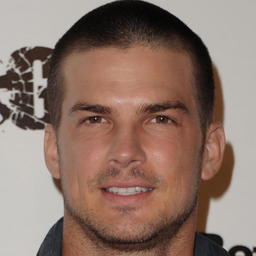}
    \end{subfigure} & 
    \begin{subfigure}{0.14\textwidth}
        \includegraphics[width=\textwidth]{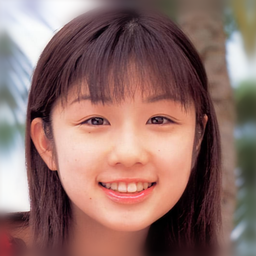}
    \end{subfigure} \\

    \rotatebox{90}{\hskip 1em TIP-IM~\cite{yang2021towards}} & 
    \begin{subfigure}{0.14\textwidth}
        \includegraphics[width=\textwidth]{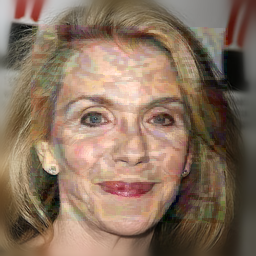}
    \end{subfigure} & 
    \begin{subfigure}{0.14\textwidth}
        \includegraphics[width=\textwidth]{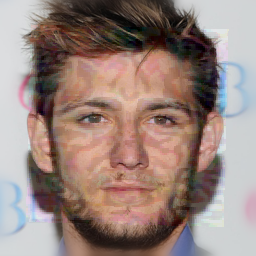}
    \end{subfigure} & 
    \begin{subfigure}{0.14\textwidth}
        \includegraphics[width=\textwidth]{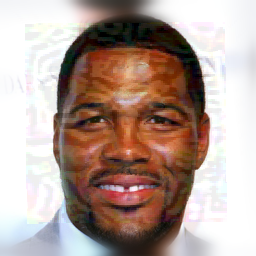}
    \end{subfigure} & 
    \begin{subfigure}{0.14\textwidth}
        \includegraphics[width=\textwidth]{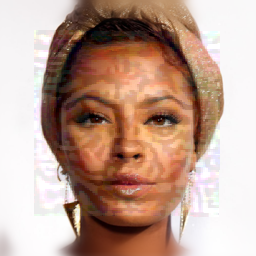}
    \end{subfigure} & 
    \begin{subfigure}{0.14\textwidth}
        \includegraphics[width=\textwidth]{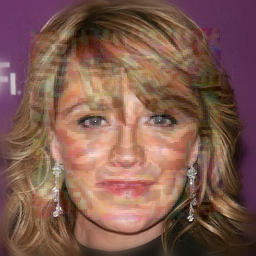}
    \end{subfigure} & 
    \begin{subfigure}{0.14\textwidth}
        \includegraphics[width=\textwidth]{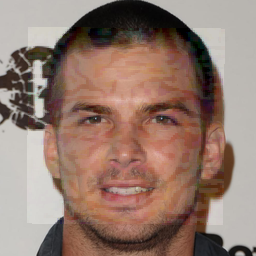}
    \end{subfigure} & 
    \begin{subfigure}{0.14\textwidth}
        \includegraphics[width=\textwidth]{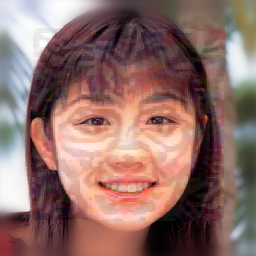}
    \end{subfigure} \\

    \rotatebox{90}{\hskip 0.5em AMT-GAN~\cite{hu2022protecting}} & 
        \begin{subfigure}{0.14\textwidth}
        \includegraphics[width=\textwidth]{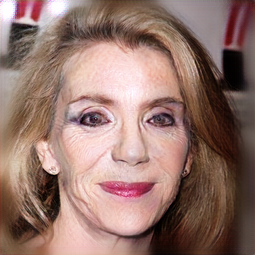}
    \end{subfigure} & 
    \begin{subfigure}{0.14\textwidth}
        \includegraphics[width=\textwidth]{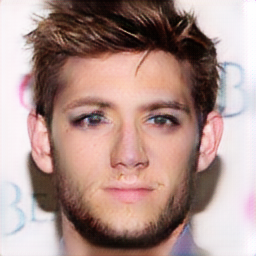}
    \end{subfigure} & 
    \begin{subfigure}{0.14\textwidth}
        \includegraphics[width=\textwidth]{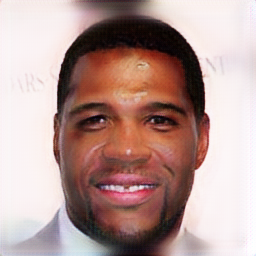}
    \end{subfigure} & 
    \begin{subfigure}{0.14\textwidth}
        \includegraphics[width=\textwidth]{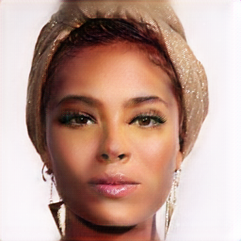}
    \end{subfigure} & 
    \begin{subfigure}{0.14\textwidth}
        \includegraphics[width=\textwidth]{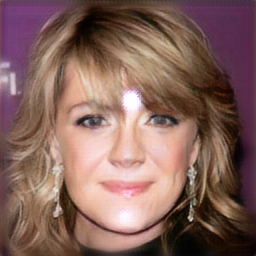}
    \end{subfigure} & 
    \begin{subfigure}{0.14\textwidth}
        \includegraphics[width=\textwidth]{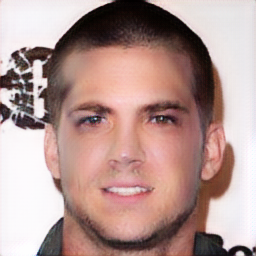}
    \end{subfigure} & 
    \begin{subfigure}{0.14\textwidth}
        \includegraphics[width=\textwidth]{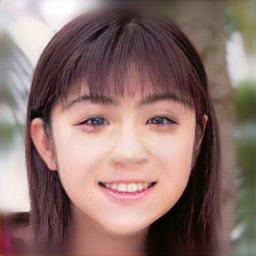}
    \end{subfigure} \\
    \rotatebox{90}{\hskip 0.5em \textbf{DiffProtect-fast}} & 
        \begin{subfigure}{0.14\textwidth}
        \includegraphics[width=\textwidth]{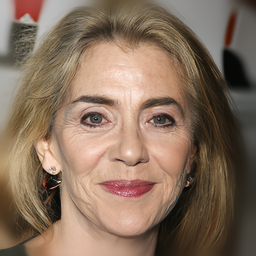}
    \end{subfigure} & 
    \begin{subfigure}{0.14\textwidth}
        \includegraphics[width=\textwidth]{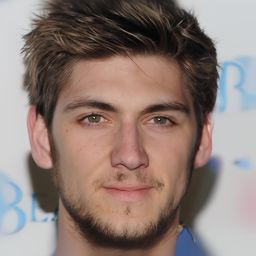}
    \end{subfigure} & 
    \begin{subfigure}{0.14\textwidth}
        \includegraphics[width=\textwidth]{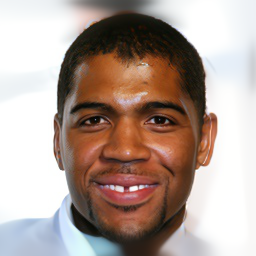}
    \end{subfigure} & 
    \begin{subfigure}{0.14\textwidth}
        \includegraphics[width=\textwidth]{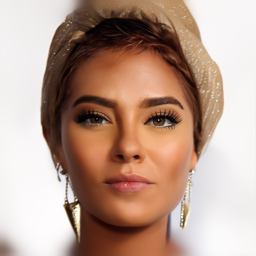}
    \end{subfigure} & 
    \begin{subfigure}{0.14\textwidth}
        \includegraphics[width=\textwidth]{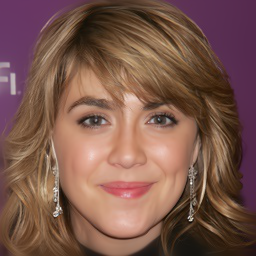}
    \end{subfigure} & 
    \begin{subfigure}{0.14\textwidth}
        \includegraphics[width=\textwidth]{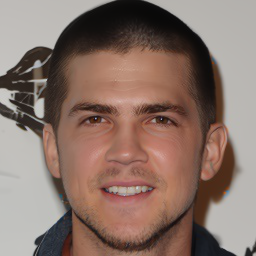}
    \end{subfigure} & 
    \begin{subfigure}{0.14\textwidth}
        \includegraphics[width=\textwidth]{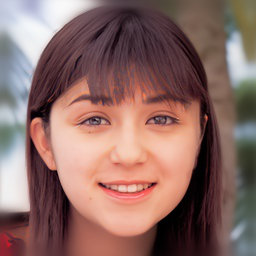}
    \end{subfigure} \\

    \rotatebox{90}{\hskip 1.5em \textbf{DiffProtect}} & 
        \begin{subfigure}{0.14\textwidth}
        \includegraphics[width=\textwidth]{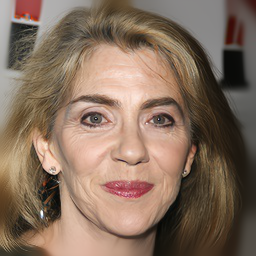}
    \end{subfigure} & 
    \begin{subfigure}{0.14\textwidth}
        \includegraphics[width=\textwidth]{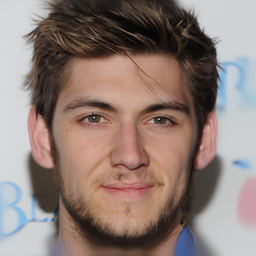}
    \end{subfigure} & 
    \begin{subfigure}{0.14\textwidth}
        \includegraphics[width=\textwidth]{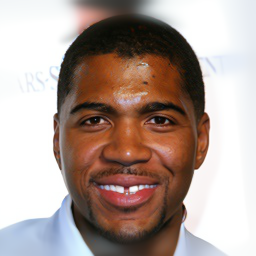}
    \end{subfigure} & 
    \begin{subfigure}{0.14\textwidth}
        \includegraphics[width=\textwidth]{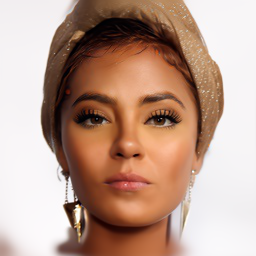}
    \end{subfigure} & 
    \begin{subfigure}{0.14\textwidth}
        \includegraphics[width=\textwidth]{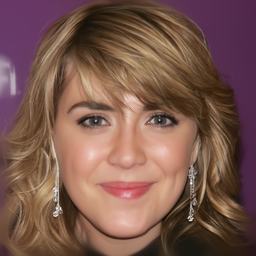}
    \end{subfigure} & 
    \begin{subfigure}{0.14\textwidth}
        \includegraphics[width=\textwidth]{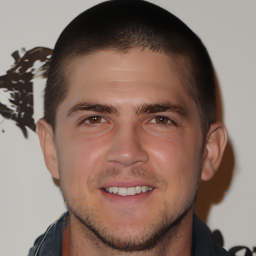}
    \end{subfigure} & 
    \begin{subfigure}{0.14\textwidth}
        \includegraphics[width=\textwidth]{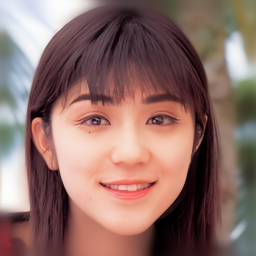}
    \end{subfigure} \\
\end{tabular}}
\caption{Visualizations of the protected face images generated by different face encryption methods on CelebA-HQ.}
\label{fig:main}
\end{figure*}
\section{Experiments}
\subsection{Experimental Settings}
\label{sec:settings}
\paragraph{Dataset} We evaluate DiffProtect on two commonly used high-quality face image datasets: CelebA-HQ~\cite{karras2017progressive} and FFHQ~\cite{karras2019style}. Following \cite{hu2022protecting}, we use a subset of 1000 face images with different identities for each dataset.
\paragraph{Attack Setting} In this paper, we focus on black-box targeted attacks following previous work~\cite{zhong2020towards, qiu2019semanticadv, hu2022protecting}.  We consider three popular face recognition models, including IR152 \cite{he2016deep}, IRSE50 \cite{hu2018squeeze}, and MobileFace \cite{deng2019arcface} as the victim models. For each model, we use the other models as the surrogate models $\gA_g$ to craft black-box attacks. 
\paragraph{Evaluation Metrics} We use Attack Success Rate (ASR)~\cite{hu2022protecting, zhong2020towards} to evaluate the attack performance: 
\begin{equation}
    \text{ASR} = \frac{1}{K}\sum_{\vect{I}}\mathbbm{I}\left(\text{cos}(h(\vect{I}_t), h(\vect{I}_\text{p}))>\tau \right)\times 100\%,
\end{equation}
where $\mathbbm{I}$ is the indicator function, $K$ is the number of face image $\vect{I}$, $\tau$ is the threshold, $\vect{I}_t$ and $\vect{I}_p$ are the target and protected face images respectively. The value of $\tau$ is set at 0.01 False Acceptance Rate (FAR) for each victim model. In addition, we use Frechet Inception Distance (FID)~\cite{heusel2017gans} to evaluate the naturalness of protected face images. 

\paragraph{Implementation Details} The architecture of DiffProtect is mainly based on Diffusion Autoencoder (DiffAE)~\cite{diffae}. We use the DiffAE model trained on the FFHQ dataset with $256\times 256$ image resolutions. The DiffAE model is fixed during attack generation. For face semantics regularization, we use a pretrained  BiSeNet \cite{yu2018bisenet} trained on the CelebAMask-HQ dataset \cite{CelebAMask-HQ} as the face parsing network $\gM$. We set $N=50$, $\gamma=0.03$, $\eta=2\cdot \frac{\gamma}{N}$ and $\lambda=0$ by default. We set $T=100$ for conditional DDIM encoding and final decoding, but use five decoding steps for reconstructing $\vect{I}_{p}^{(i)}$ during attack (\cref{eq:Ipi}) in order to save time and memory. For DiffProtect-fast, we set $t_o=60$. All experiments were run on a server with 8 Nvidia A5000 GPUs.

\begin{figure}[t]
\centering
 \setlength{\tabcolsep}{0.5mm}
 \scalebox{1}{
\begin{tabular}[b]{ccc}
    \rotatebox{90}{\hskip 3em \small{IRSE50} } &
    \begin{subfigure}{0.23\textwidth}
        \includegraphics[width=\textwidth]{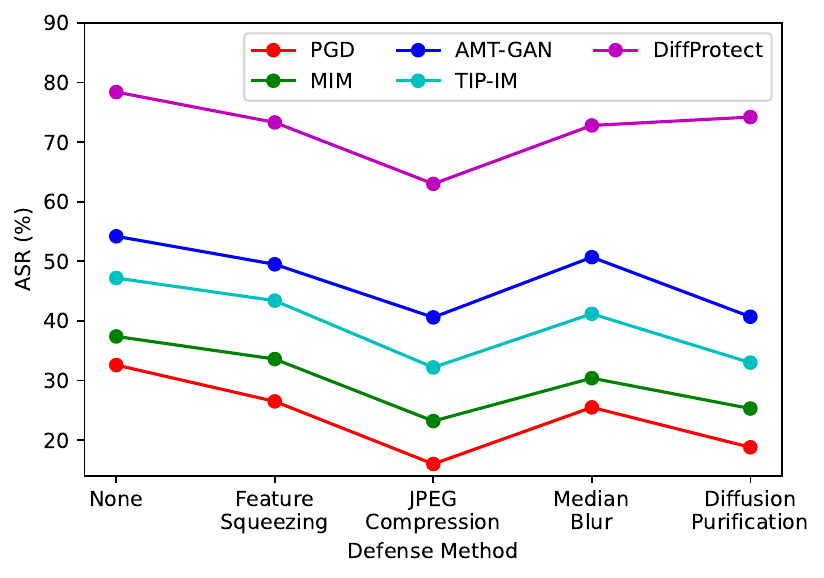}
    \end{subfigure} & 
    \begin{subfigure}{0.23\textwidth}
        \includegraphics[width=\textwidth]{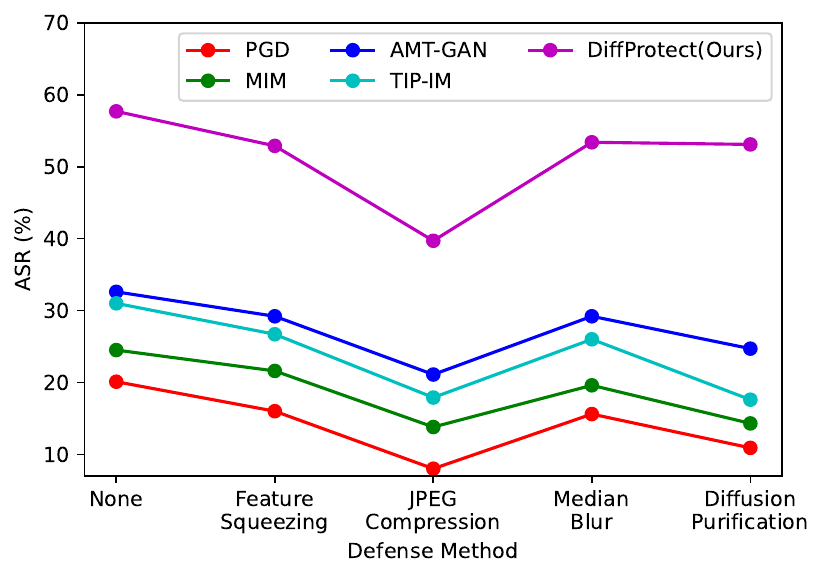}
    \end{subfigure} \\
    
    \rotatebox{90}{\hskip 3em \small{IR152}} &
    \begin{subfigure}{0.23\textwidth}
        \includegraphics[width=\textwidth]{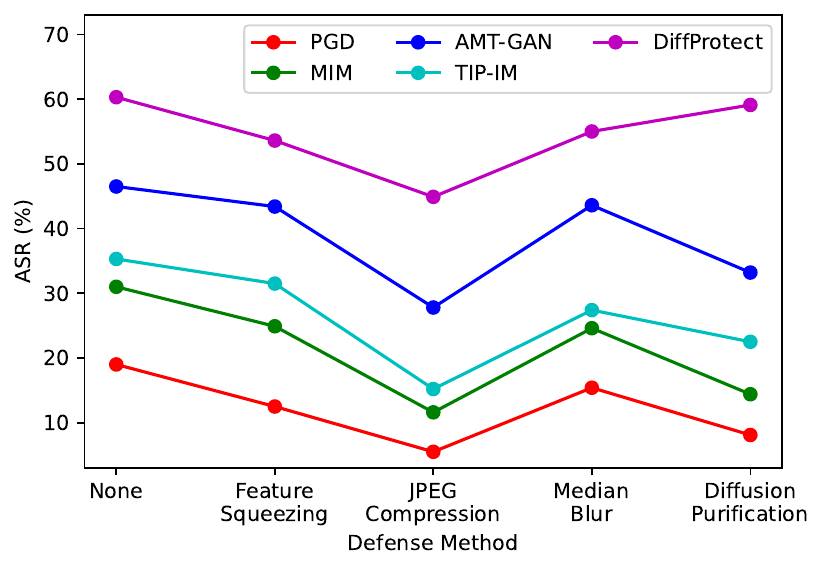}
    \end{subfigure} & 
    \begin{subfigure}{0.23\textwidth}
        \includegraphics[width=\textwidth]{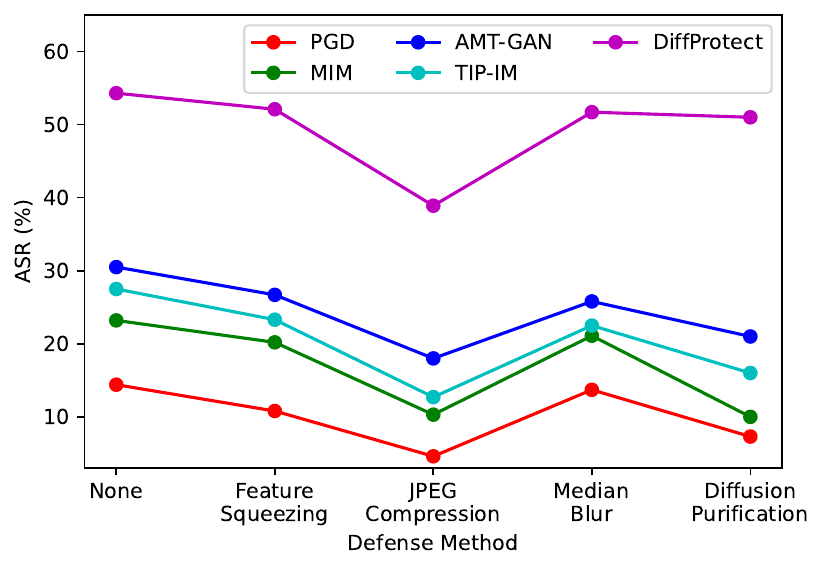}
        
    \end{subfigure} \\

    \rotatebox{90}{\hskip 4em \small{MobileFace}} &
    \begin{subfigure}{0.23\textwidth}
        \includegraphics[width=\textwidth]{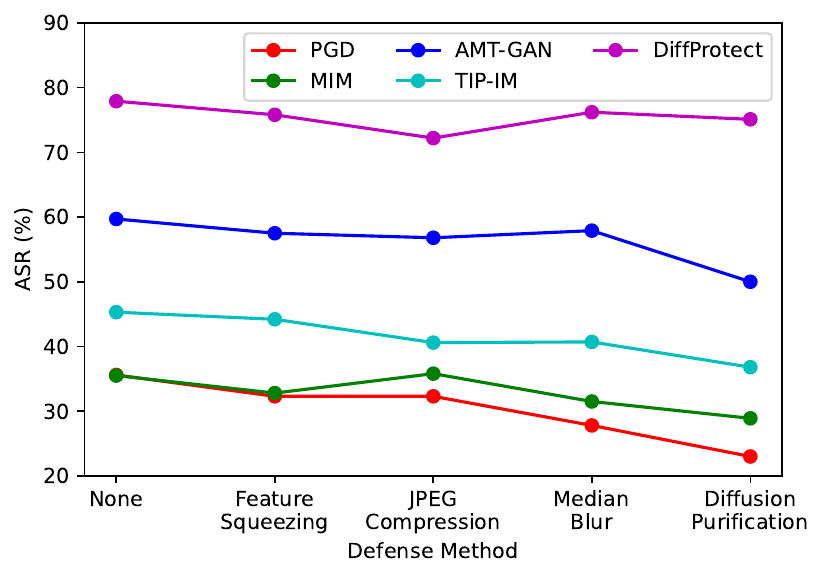}
        \caption{CelebA-HQ}
    \end{subfigure} &
    \begin{subfigure}{0.23\textwidth}
        \includegraphics[width=\textwidth]{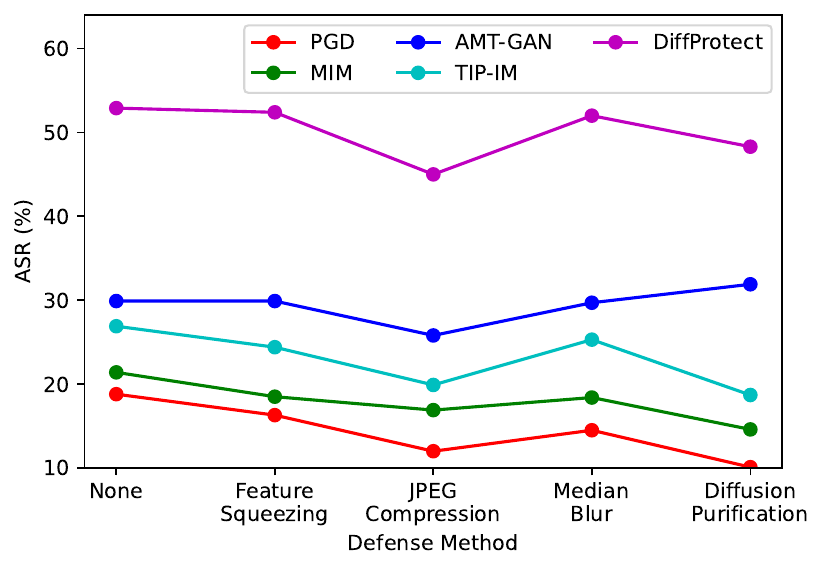}
        \caption{FFHQ}
    \end{subfigure} \\
\end{tabular}}
\caption{ASR under various defense methods.}
\label{fig:defense}
\end{figure}
\subsection{Comparison with the state-of-the-art}
\paragraph{Main Results} We compare DiffProtect with state-of-the-art face encryption methods, including PGD~\cite{madry2017towards}, MIM~\cite{dong2018boosting}, TIP-IM~\cite{yang2021towards}, and AMT-GAN~\cite{hu2022protecting}. The implementation details of baseline methods can be found in the supplementary materials.~\Cref{tab:asr_main} reports the quantitative results on the CelebA-HQ and FFHQ datasets. For ASR, DiffProtect significantly outperforms previous methods by a large margin. In addition, DiffProtect achieves the lowest FID scores, which indicates that the encrypted images generated by DiffProtect are the most natural. We show some examples of protected face images in~\cref{fig:main} with $\lambda=0$. We can observe that DiffProtect produces good-looking protected images with natural and inconspicuous changes to the input images, such as slight changes in facial expressions. It works well across genders, ages, and races, and in some cases even makes the images look more attractive. Compared to TIP-IM, the protected face images generated by DiffProtect have no obvious noise pattern as we only perturb the semantic codes and generate the images through a conditional DDIM. Compared to AMT-GAN, DiffProtect can better preserve image styles and details and does not require training a target-specific model for each identity. Our accelerated method DiffProtect-fast also achieves higher ASRs and lower FID compared to the baselines. In addition, we can observe that DiffProtect and DiffProtect-fast achieve similar visual quality, while DiffProtect-fast requires 50\% less computation time (more details in the supplementary material).

\paragraph{ASR under Defenses} We further evaluate the effectiveness of DiffProtect against four adversarial defense methods, including feature squeezing~\cite{xu2017feature}, median blurring~\cite{blur}, JPEG compression~\cite{jpeg}, and a state-of-the-art diffusion-based defense DiffPure~\cite{nie2022DiffPure}. The implementation details of the defenses can be found in the supplementary materials. From~\cref{fig:defense} we can see that DiffProtect still achieves higher ASR under various defenses. In addition, DiffProtect is much more resilient to the DiffPure defense compared to baseline methods since it generates semantically meaningful perturbations to the input images that are hard to remove during the diffusion denoising process of DiffPure.

\label{sec:acceleration}
\subsection{Ablation Studies}
For ablation studies, we use the CelebA dataset and MobileFace as the victim model. We set $N=10$, $\gamma=0.02$, and $T=50$. The other settings are the same as~\cref{sec:settings}.
\begin{table}
\centering
\setlength{\tabcolsep}{1mm}
\scalebox{0.95}{\begin{tabular}{c|c|ccc}
\toprule[1pt]
{Dataset}   & {Model}   & IRSE50 & IR152 & MobileFace        \\
\hline
\multirow{2}{*}{CelebA-HQ} & GAN    & 33.0     & 16.5  & 41.1       \\
          & Diff. & \textbf{60.7}\small{ (+27.7)}   & \textbf{37.5}\small{ (+21.0)}  & \textbf{62.0}\small{ (+20.6)}         \\
\hline
\multirow{2}{*}{FFHQ}      & GAN    & 21.3   & 20.0    & 21.2        \\
          & Diff. & \textbf{36.5}\small{ (+15.2)}   & \textbf{33.1}\small{ (+13.1)}  & \textbf{34.0}\small{ (+12.8)}         \\
\bottomrule[1pt]
\end{tabular}}
\caption{ASR (\%) of GAN~\cite{wang2021HFGI} and Diffusion~\cite{diffae} model.}
\label{tab:gan}
\end{table}

\begin{figure}[t]
    \centering
    \small
    \setlength{\tabcolsep}{1mm}
    \scalebox{0.9}{
\begin{tabular}[b]{cccc} 
    \begin{subfigure}{0.16\textwidth}
        \includegraphics[width=\textwidth]{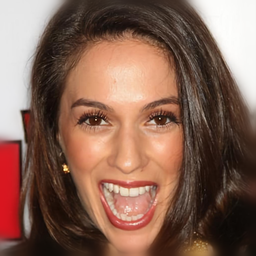}
    \end{subfigure} & 
    \begin{subfigure}{0.16\textwidth}
        \includegraphics[width=\textwidth]{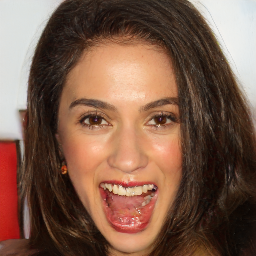}
    \end{subfigure} & 
    \begin{subfigure}{0.16\textwidth}
        \includegraphics[width=\textwidth]{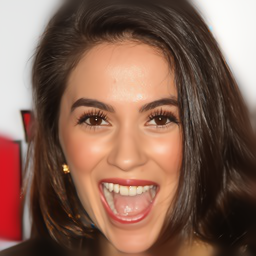}
    \end{subfigure} \\
    
    \begin{subfigure}{0.16\textwidth}
        \includegraphics[width=\textwidth]{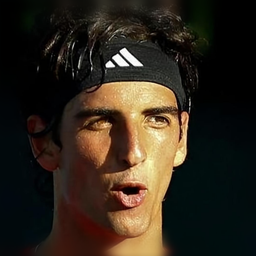}
        \caption{Input image}
    \end{subfigure} & 
    \begin{subfigure}{0.16\textwidth}
        \includegraphics[width=\textwidth]{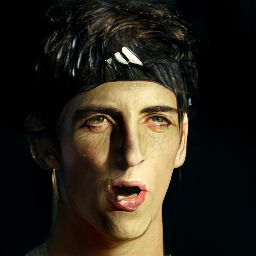}
        \caption{GAN output.}
    \end{subfigure} & 
    \begin{subfigure}{0.16\textwidth}
        \includegraphics[width=\textwidth]{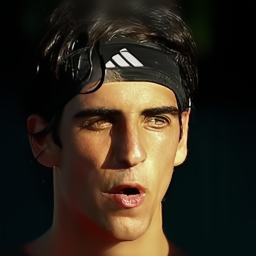}
        \caption{Diffusion output.}
    \end{subfigure} \\
\end{tabular}}
\caption{Visualizations of adversarial images generated by GAN~\cite{wang2021HFGI} and diffusion~\cite{diffae} models on CelebA-HQ.  }
\label{fig:gan}
\end{figure}
\begin{figure*}[htbp]
\captionsetup[subfigure]{labelformat=empty}
    \centering
    \small
    \setlength{\tabcolsep}{0.5mm}{
\begin{tabular}[b]{cccccccc}
    Input Image & $\gamma=0.005$ & $\gamma=0.01$ & $\gamma=0.02$ & $\gamma=0.03$ & $\gamma=0.05$ & $\gamma=0.1$ & Target Image\\
    \begin{subfigure}{0.12\textwidth}
        \includegraphics[width=\textwidth]{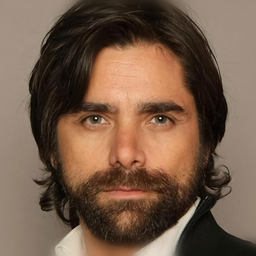}
    \end{subfigure} & 
    \begin{subfigure}{0.12\textwidth}
        \includegraphics[width=\textwidth]{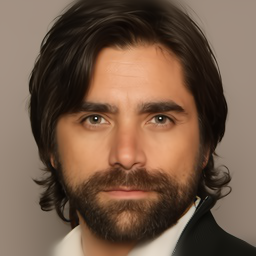}
    \end{subfigure} & 
    \begin{subfigure}{0.12\textwidth}
        \includegraphics[width=\textwidth]{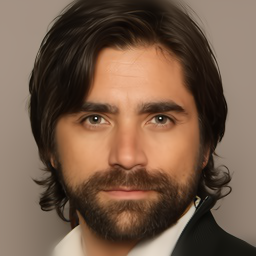}
    \end{subfigure} & 
    \begin{subfigure}{0.12\textwidth}
        \includegraphics[width=\textwidth]{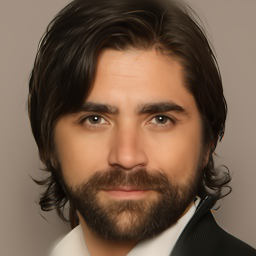}
    \end{subfigure} & 
    \begin{subfigure}{0.12\textwidth}
        \includegraphics[width=\textwidth]{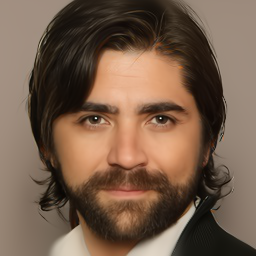}
    \end{subfigure} & 
    \begin{subfigure}{0.12\textwidth}
        \includegraphics[width=\textwidth]{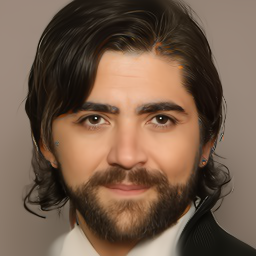}
    \end{subfigure} & 
    \begin{subfigure}{0.12\textwidth}
        \includegraphics[width=\textwidth]{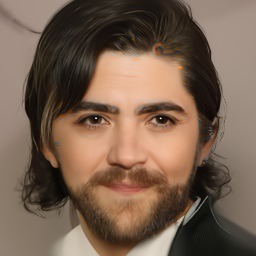}
    \end{subfigure} &
    \begin{subfigure}{0.12\textwidth}
        \includegraphics[width=\textwidth]{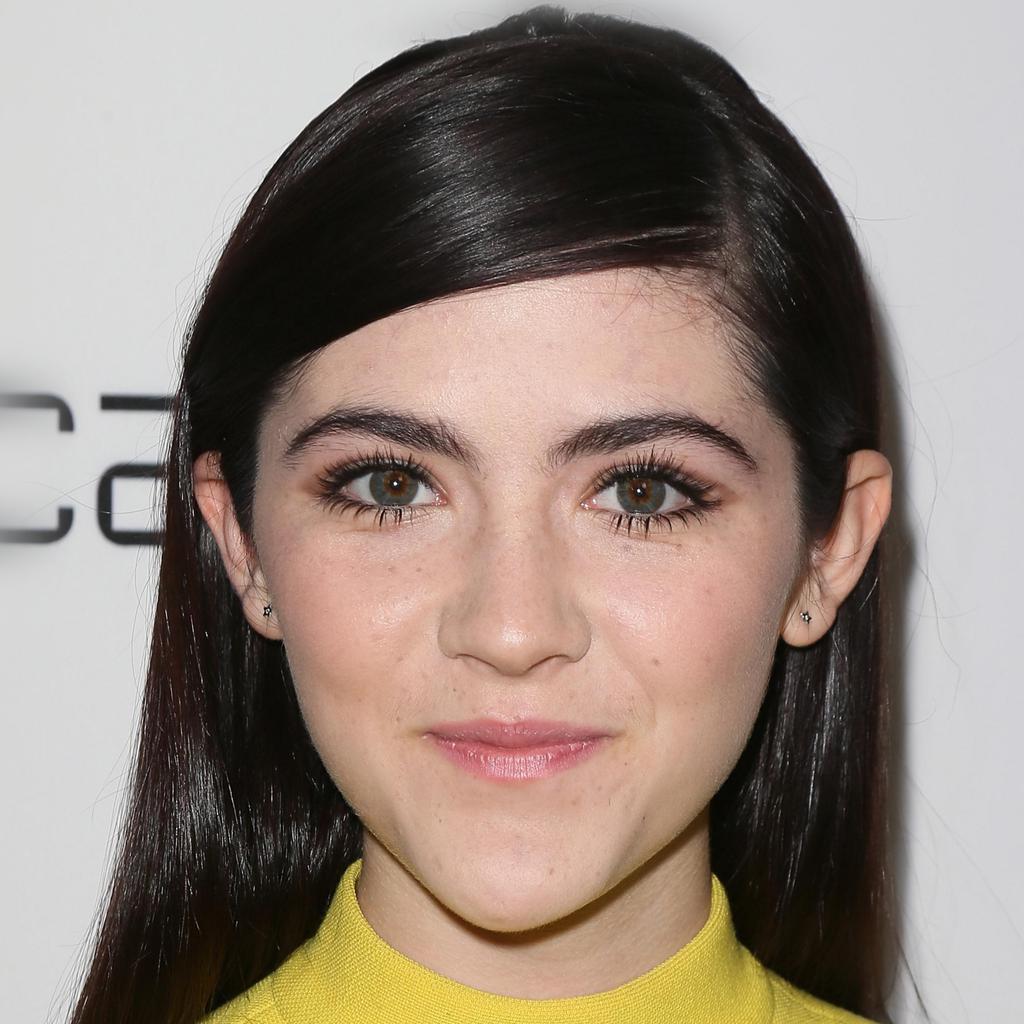}
    \end{subfigure}\\
    
    \begin{subfigure}{0.12\textwidth}
        \includegraphics[width=\textwidth]{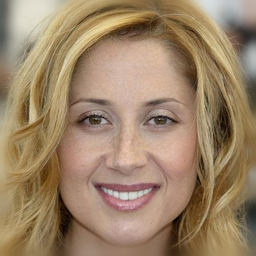}
    \end{subfigure} & 
    \begin{subfigure}{0.12\textwidth}
        \includegraphics[width=\textwidth]{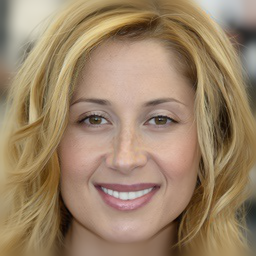}
    \end{subfigure} & 
    \begin{subfigure}{0.12\textwidth}
        \includegraphics[width=\textwidth]{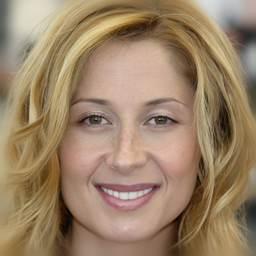}
    \end{subfigure} & 
    \begin{subfigure}{0.12\textwidth}
        \includegraphics[width=\textwidth]{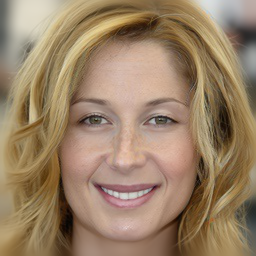}
    \end{subfigure} & 
    \begin{subfigure}{0.12\textwidth}
        \includegraphics[width=\textwidth]{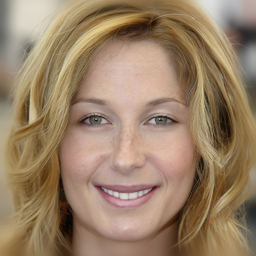}
    \end{subfigure} & 
    \begin{subfigure}{0.12\textwidth}
        \includegraphics[width=\textwidth]{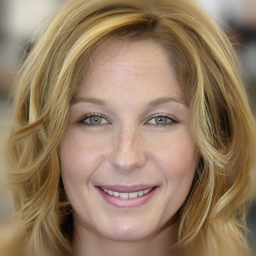}
    \end{subfigure} & 
    \begin{subfigure}{0.12\textwidth}
        \includegraphics[width=\textwidth]{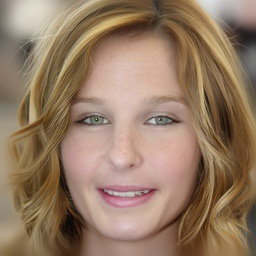}
    \end{subfigure} & 
    \begin{subfigure}{0.12\textwidth}
        \includegraphics[width=\textwidth]{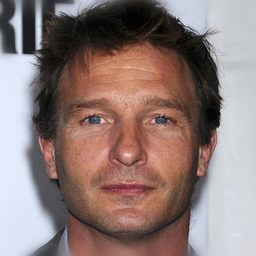}
    \end{subfigure}\\
    \scalebox{1.2}{$\xrightarrow[]{\ \ \ \gamma\ \text{increasing}\ \ \ }$} & \asr{25.0\%}, \fid{28.5} & \asr{37.4\%}, \fid{30.23} & \asr{60.5\%}, \fid{33.42} & \asr{75.4\%}, \fid{36.42} & \asr{88.4\%}, \fid{42.53} & \asr{97.7\%}, \fid{57.25} \\
\end{tabular}}
\caption{Visual comparisons of different attack budget $\gamma$ and the corresponding \asr{ASR} and \fid{FID}.}
\label{fig:budget}
\end{figure*}
\paragraph{GAN \textit{vs.} Diffusion model} To have a fair comparison between GAN-based and Diffusion-based methods, we re-implement our method using a GAN-based autoencoder~\cite{wang2021HFGI} where we also iteratively optimize the GAN latent code using the same formulation. The implementation details can be found in the supplementary materials. The results are summarized in~\Cref{tab:gan}. We can observe that the diffusion-based method has much higher ASRs than the GAN-based method. From~\cref{fig:gan} we can observe that the image quality of Diffusion model output is also visually better than the GAN-based method. These results confirm our intuition that a better generative model like a Diffusion model can help to improve both attack performance and image quality. 

\paragraph{Effect of Attack Budget} We show the effect of attack budget $\gamma$ in~\cref{fig:budget}. We can observe that as $\gamma$ increases, the ASR also increases; when $\gamma=0$, we have $\text{ASR}=97.7\%$. However, the FID also goes up as $\gamma$ increases, which indicates that the image quality becomes worse. In addition, a large $\gamma$ can cause the protected image to lose the visual identity of the input image. In practice, we recommend setting $\gamma \leq 0.03$ such that the output images have high visual quality and preserve the identity of the input images, which is confirmed by a user study using Amazon Mechanical Turk (see the supplementary material).

\begin{figure}[t]
\captionsetup[subfigure]{labelformat=empty}
    \centering
    \includegraphics[width=0.45\textwidth]{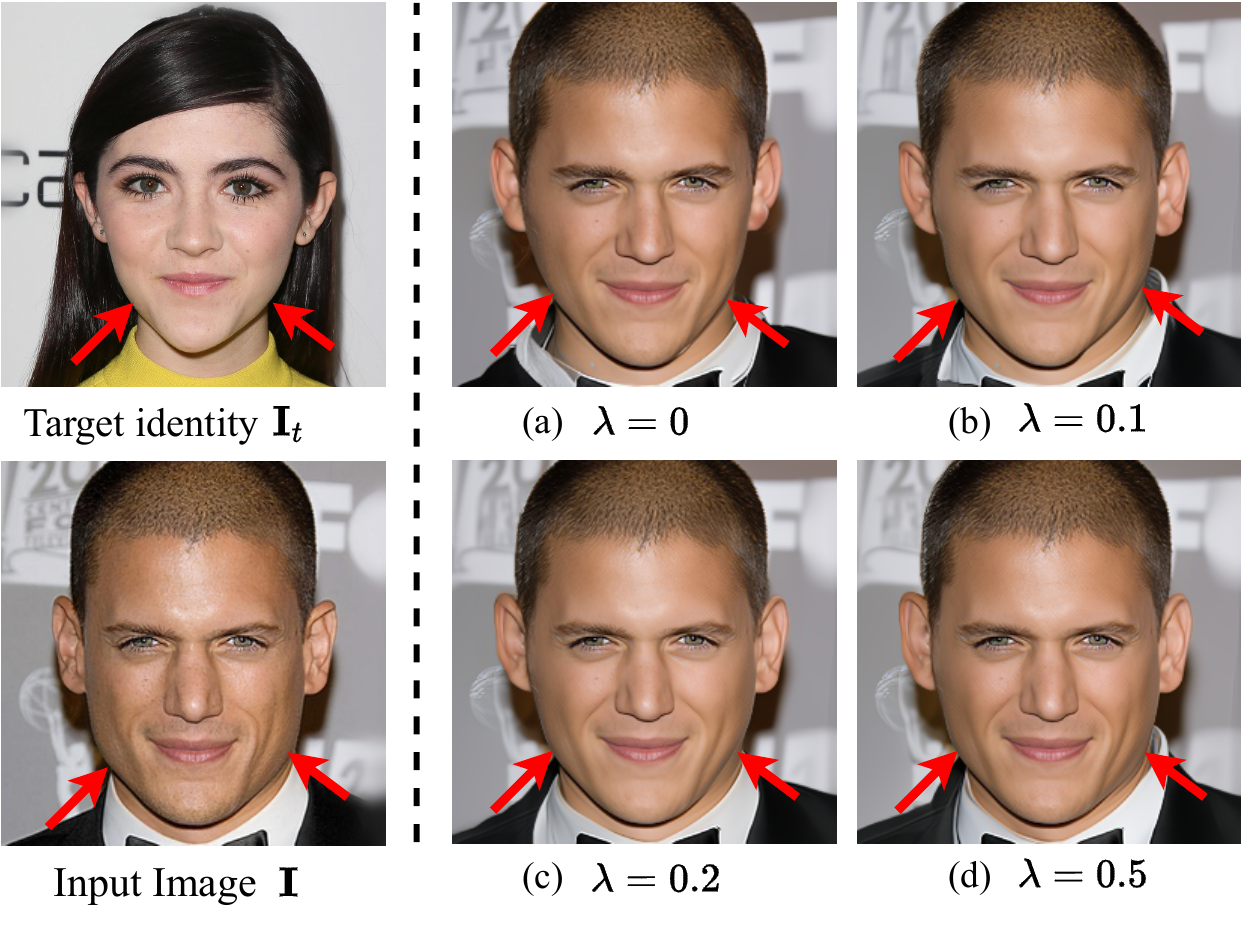}
\caption{Effect of Face Semantics Regularization. (a)-(d): protected image $\vect{I}_p$ generated with different $\lambda$ in~\cref{eq:attack_total}. Increasing $\lambda$ gradually changes the face shape from a heart shape (similar to $\textbf{I}_t$) to a square shape (similar to $\textbf{I}$).}
\label{fig:face_parsing}
\end{figure}

\paragraph{Effect of Face Semantics Regularization} We show the effect of face semantics regularization in~\cref{fig:face_parsing}. Without the face semantics regularization term ($\lambda=0$), the protected image can have different local facial features compared to the input image to match the characteristics of the target image, such as the face shape in~\cref{fig:face_parsing} (a). Increasing the value of $\lambda$ helps to preserve the face features of the input images. Quantitatively, from~\Cref{tab:parsing} we can observe that increasing $\lambda$ helps to reduce FID, which indicates better image quality but also decreases ASR. In practice, a user can choose a value of $\lambda$ as well as the attack budget $\gamma$ depending on their own preference.
\begin{table}[t]
\centering
\begin{tabular}{c|cccccc}
\toprule
$\lambda$ & 0     & 0.01  & 0.05  & 0.1   & 0.2   & 0.5   \\
\midrule
ASR (\%) $\uparrow$   & 60.5  & 60.3  & 58.9  & 56.3  & 53.5  & 46.1  \\
FID $\downarrow$   & 33.4 & 33.3 & 32.8 & 32.4 & 31.9 & 31.2  \\
\bottomrule
\end{tabular}
\caption{Quantitative results of face semantics regularization with different $\lambda$ in~\cref{eq:attack_total}.}
\label{tab:parsing}
\end{table}



\begin{figure}[t]
    \centering
    \includegraphics[width=0.4\textwidth]{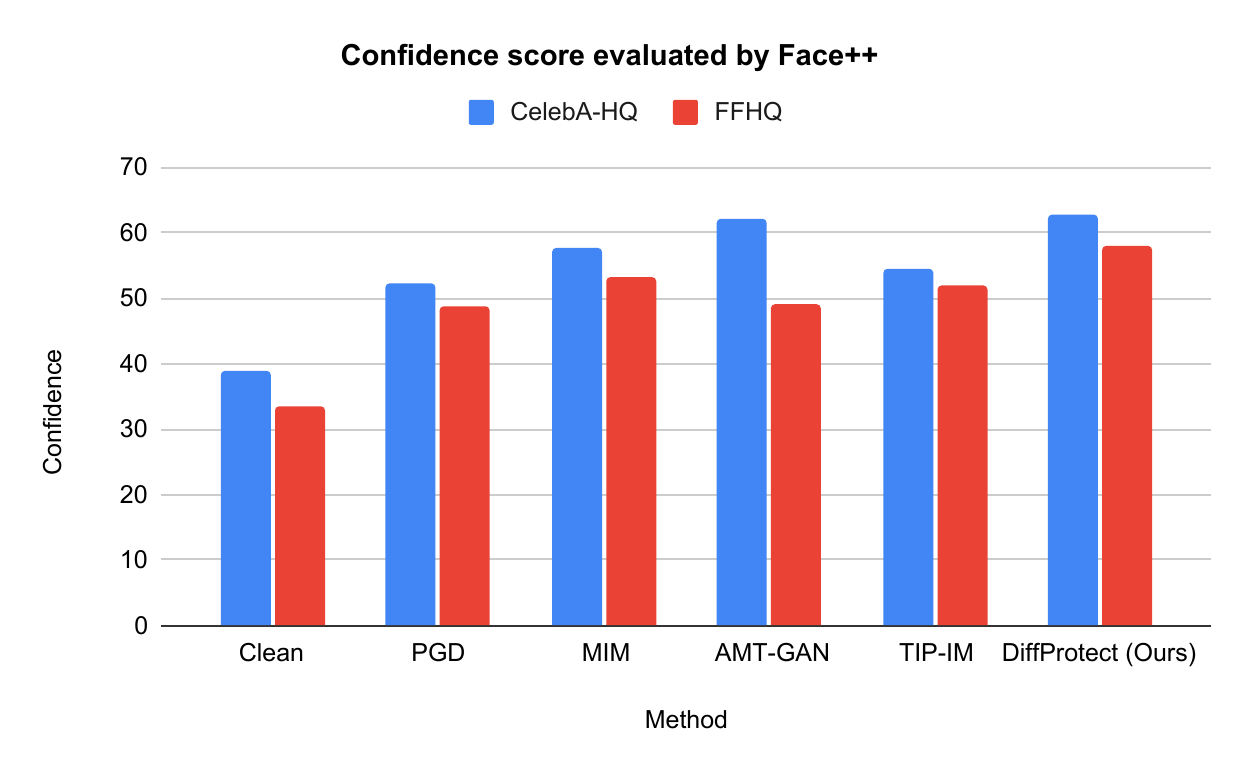}
     \caption{Confidence scores of different methods evaluated by commercial API Face++.}
     \label{fig:face_api}
\end{figure}

\subsection{Evaluation on Commercial API} We further evaluate the effectiveness of AGE-FTM in the real world using a commercial FR API Face++\footnote{\url{https://www.faceplusplus.com/}}, where we compute the confidence scores between the protected images and the target identity image. DiffProtect achieves average confidence scores of 62.8\% and 58.03\% for CelebA-HQ and FFHQ respectively, which are higher than baseline methods. This shows that the proposed method has a satisfactory attack ability even for the commercial API while achieving good image quality. 

\section{Conclusion}
In this work, we propose DiffProtect a novel diffusion-based method for facial
privacy protection. It generates semantically meaningful perturbations to the input images and produces adversarial images of high visual quality. Our extensive experiments on CelebA-HQ and FFHQ demonstrate that DiffProtect significantly outperforms previous state-of-the-art methods for both attack performance and image quality.

{\small
\bibliographystyle{ieee_fullname}
\bibliography{ref}
}

\newpage
\appendix
\section*{Supplementary Material}
\begin{table*}[t]
\centering
\setlength{\tabcolsep}{0.2mm}
\scalebox{0.94}{\setlength{\tabcolsep}{2.0mm}\begin{tabular}{l|ccc|c|c|ccc|c|cc}
\toprule[1pt]
\multirow{3}{*}{Methods}     & \multicolumn{5}{c|}{CelebA-HQ}  & \multicolumn{5}{c}{FFHQ}  \\
\cline{2-11}
 & \multicolumn{3}{c|}{ASR (\%) $\uparrow$} & \multirow{2}{*}{FID $\downarrow$} & \multirow{2}{*}{Time (s) $\downarrow$} & \multicolumn{3}{c|}{ASR (\%) $\uparrow$} & \multirow{2}{*}{FID $\downarrow$} & \multirow{2}{*}{Time (s) $\downarrow$} \\
 
\cline{2-4} \cline{7-9}
             &  IRSE50    & IR152 & MobileFace &  & & IRSE50 & IR152 & MobileFace &  \\
\hline
DiffProtect-fast & {68.4}      & {49.8}  & {72.1}     &  \textbf{26.7} & \textbf{18.9} & {50.8}   & {47.6}  & {47.0} &   26.4 & \textbf{18.3} \\
DiffProtect   & \textbf{78.4}      & \textbf{60.3}  & \textbf{77.9}  & {27.6} & 36.4  & \textbf{57.7}   & \textbf{54.3}  & \textbf{52.9} & \textbf{26.1} & 35.9 \\
\bottomrule[1pt]
\end{tabular}}
\caption{Comparison between DiffProtect and DiffProtect-fast.}
\label{tab:fast}
\end{table*}
\section{DiffProtect-fast}
\subsection{Effect of $t_0$} 
\begin{figure}[h]
    \centering
    \includegraphics[width=0.45\textwidth]{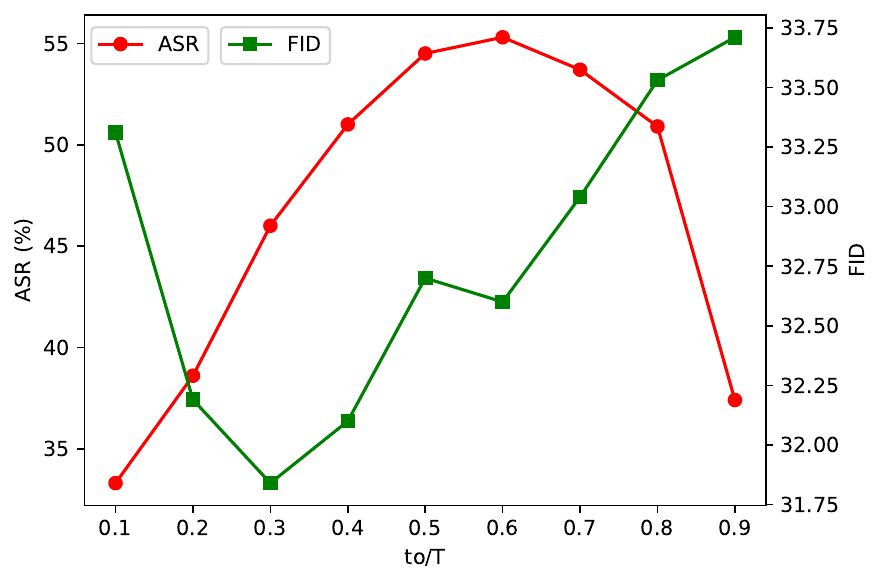}
     \caption{Effect of $t_0/T$ ratio for DiffProtect-fast.}
     \label{fig:fast}
\end{figure}

In DiffProtect-fast, we run one generative step from the time stamp $t_0$ to compute $\vect{I}_{p}^{(i)}$, instead of running the whole generative process from $t=T$ to $t=0$ to accelerate attack generation. We show the effect of the choice of $t_0$ in~\cref{fig:fast}, where we use the same settings as in the ablation studies. We can observe that when we increase $t_0$, ASR first increases and then decreases, while FID first decreases and then increases, which indicates the image quality also first increases and then decreases. This shows that choosing $t_0$ at the early or late stage of the generative process is less effective. We set $t_0/T = 0.6$, \ie, $t_0=0.6\cdot T$,  as this ratio achieves the highest ASR and is also a local minimum for FID.

\subsection{Comparison with DiffProtect}
We provide quantitative comparisons between DiffProtect and DiffProtect-fast in~\Cref{tab:fast}. Compared to DiffProtect, DiffProtect-fast achieves lower ASRs, and similar FIDs, but requires much less computation time. Note that DiffProtect-fast also outperforms the baseline methods as shown in Table {\color{red}1} of the main paper. These results demonstrate that the attack acceleration strategy is very effective and we can use DiffProtect-fast to achieve a trade-off between computation time and attack success rate. In addition, by comparing the last two rows of~\cref{fig:supp,fig:supp_2}, we can observe that the protected images generated by DiffProtect and DiffProtect-fast are visually similar.

\section{Additional Implementation Details}
\subsection{Baselines}
For $l_\infty$ attacks such as PGD~\cite{madry2017towards}, MIM~\cite{dong2018boosting} and TIP-IM~\cite{yang2021towards}, the attack budget is set to be $16/255$ with $100$ attack iterations. For MIM, the decay factor $\mu$ is set to $1.0$. For TIP-IM, the gamma value for the MMD loss is set to $0$ by default. For AMT-GAN~\cite{hu2022protecting}, we use the default setting. 
\subsection{Defense methods}
We evaluated our methods with four defense methods: feature squeezing \cite{gu2019detecting}, JPEG compression \cite{dziugaite2016study}, median blur \cite{li2017adversarial}, and adversarial purification \cite{nie2022DiffPure}. For feature squeezing, the bit depth is set to $3$. For JPEG compression, the quality factor is set to $5$. For the median blur, the kernel size is set to $7$. For adversarial purification, we set the diffusion timestep to be $0.0075$. 
\subsection{GAN-based attack}
To have a fair comparison between GAN-based and Diffusion-based methods,
we re-implement our method using an HFGI \cite{wang2021HFGI} model, a state-of-the-art method that uses a high-fidelity GAN and can invert and edit high-resolution face images effectively. HFGI consists of an encoder $E$ to encode the images and uses a generator $G$ to edit the high-resolution face images. To craft the GAN-based attack, we first pass the image to the encoder to get the latent code $z=E(I)$. Then we perturb $z$ by minimizing Eq. ({\color{red}15}) of the main paper to obtain the adversarial latent code $z^{adv}$. Finally, we decode $z^{adv}$ to an adversarial image using the generator $G$. We set the attack budget to $0.1$, with step size $0.02$ in $10$ iterations. 

\section{Additional Experimental Results}
We show the effect of attack iterations in~\Cref{tab:iteration}, where we change the number of attack iterations $N$ and the other settings are the same as in the ablation studies. We can observe that as $N$ increases, ASR also increases while FID stays relatively the same. We set $N=50$ in our main experiments. 

\begin{table}
\centering
\begin{tabular}{c|ccccc}
\toprule
$N$ & 2 & 5 & 10 & 20 & 50 \\ 
\midrule
ASR (\%) $\uparrow$ & 47.3 & 57.2 & 60.5 & 61.9 & 63.1 \\
FID $\downarrow$ & 32.4 & 33.3 & 33.4 & 33.6 &  33.7 \\
\bottomrule 
\end{tabular}
\caption{Effect of attack iteration $N$.}
\label{tab:iteration}
\end{table}

\section{User Study}
To further investigate the effect of different attack budgets for preserving facial identity, we conduct a user study using Amazon Mechanical Turk (MTurk), where we show the worker the original image $\mathbf{I}$ and the protected image $\mathbf{I}_p$ generated with different attack budget $\gamma$. For each pair of $\mathbf{I}$ and $\mathbf{I}_p$, the worker is asked to provide a rating from 1 to 10 of ``how likely the two face images belong to the same person?" with 1 being ``extremely unlikely" and 10 being ``extremely likely". We randomly selected 50 images from our CelebA-HQ test set for the user study and each pair of $\mathbf{I}$ and $\mathbf{I}_p$ is evaluated by ten MTurk workers. The average ratings are shown in~\Cref{tab:rating}. We chose $\gamma=0.03$ since it achieves a high user rating as well as a high ASR (78.4\%) and low FID (27.6). In practice, a user can choose the value of $\gamma$ themselves based on their own preferences. 

\begin{table}
\centering
\begin{tabular}{c|cccccc}
\toprule
$\gamma$ & 0.005 & 0.01 & 0.02 & 0.03 & 0.05 & 0.1  \\ \midrule
Rating $\uparrow$ & 8.9   & 8.7  & 7.9  & 7.0  & 5.5  & 4.7 \\
\bottomrule 
\end{tabular}
\caption{Identity similarity ratings of different $\gamma$.}
\label{tab:rating}
\end{table}

\section{More Visualization Results}
\begin{figure*}[t]
\captionsetup[subfigure]{labelformat=empty}
    \centering
    \small
    \setlength{\tabcolsep}{0.5mm}
\scalebox{0.98}{
\begin{tabular}[b]{cccccccc}
    \rotatebox{90}{\hskip 2.5em Input} & 
    \begin{subfigure}{0.14\textwidth}
        \includegraphics[width=\textwidth]{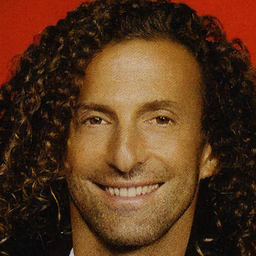}
    \end{subfigure} & 
    \begin{subfigure}{0.14\textwidth}
        \includegraphics[width=\textwidth]{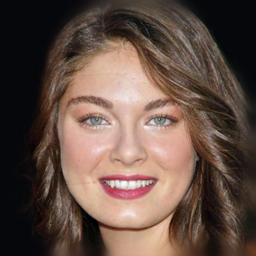}
    \end{subfigure} & 
    \begin{subfigure}{0.14\textwidth}
        \includegraphics[width=\textwidth]{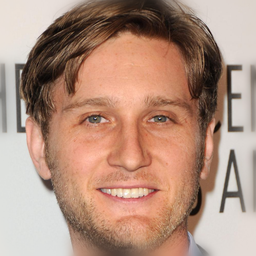}
    \end{subfigure} & 
    \begin{subfigure}{0.14\textwidth}
        \includegraphics[width=\textwidth]{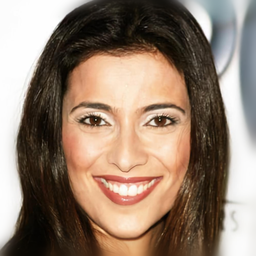}
    \end{subfigure} & 
    \begin{subfigure}{0.14\textwidth}
        \includegraphics[width=\textwidth]{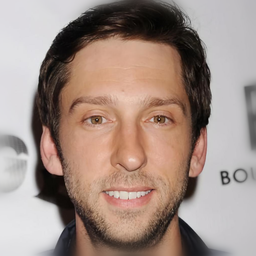}
    \end{subfigure} & 
    \begin{subfigure}{0.14\textwidth}
        \includegraphics[width=\textwidth]{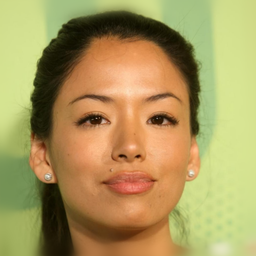}
    \end{subfigure} & 
    \begin{subfigure}{0.14\textwidth}
        \includegraphics[width=\textwidth]{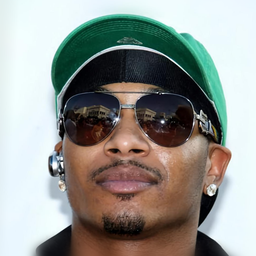}
    \end{subfigure} \\

    \rotatebox{90}{\hskip 1em TIP-IM~\cite{yang2021towards}} & 
    \begin{subfigure}{0.14\textwidth}
        \includegraphics[width=\textwidth]{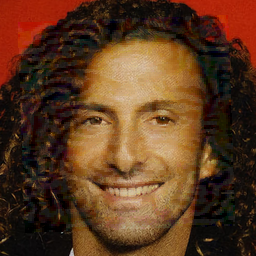}
    \end{subfigure} & 
    \begin{subfigure}{0.14\textwidth}
        \includegraphics[width=\textwidth]{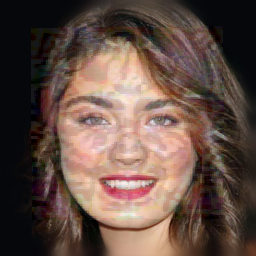}
    \end{subfigure} & 
    \begin{subfigure}{0.14\textwidth}
        \includegraphics[width=\textwidth]{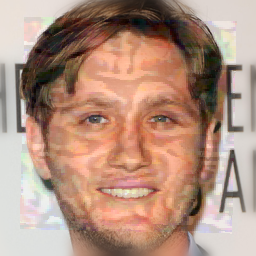}
    \end{subfigure} & 
    \begin{subfigure}{0.14\textwidth}
        \includegraphics[width=\textwidth]{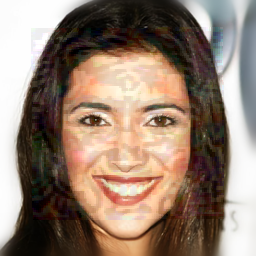}
    \end{subfigure} & 
    \begin{subfigure}{0.14\textwidth}
        \includegraphics[width=\textwidth]{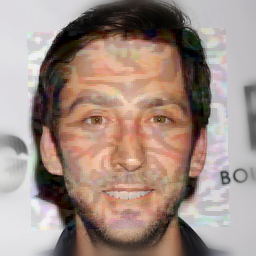}
    \end{subfigure} & 
    \begin{subfigure}{0.14\textwidth}
        \includegraphics[width=\textwidth]{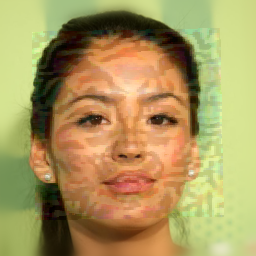}
    \end{subfigure} & 
    \begin{subfigure}{0.14\textwidth}
        \includegraphics[width=\textwidth]{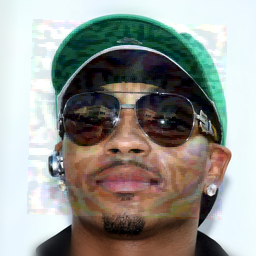}
    \end{subfigure} \\

    \rotatebox{90}{\hskip 0.5em AMT-GAN~\cite{hu2022protecting}} & 
    \begin{subfigure}{0.14\textwidth}
        \includegraphics[width=\textwidth]{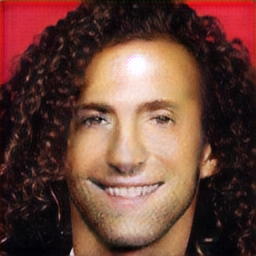}
    \end{subfigure} & 
    \begin{subfigure}{0.14\textwidth}
        \includegraphics[width=\textwidth]{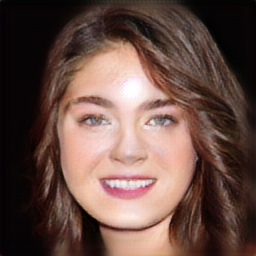}
    \end{subfigure} & 
    \begin{subfigure}{0.14\textwidth}
        \includegraphics[width=\textwidth]{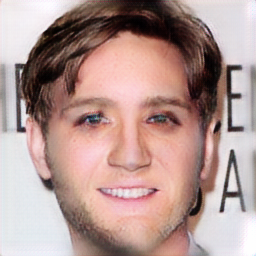}
    \end{subfigure} & 
    \begin{subfigure}{0.14\textwidth}
        \includegraphics[width=\textwidth]{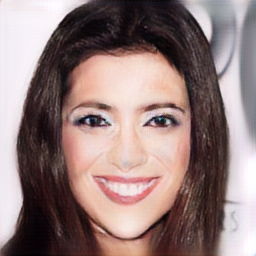}
    \end{subfigure} & 
    \begin{subfigure}{0.14\textwidth}
        \includegraphics[width=\textwidth]{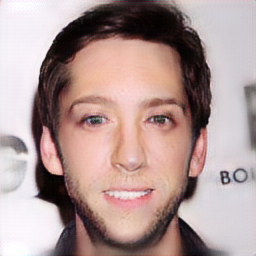}
    \end{subfigure} & 
    \begin{subfigure}{0.14\textwidth}
        \includegraphics[width=\textwidth]{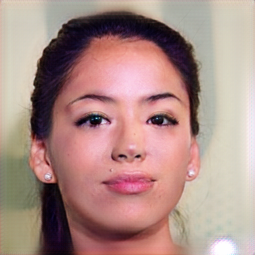}
    \end{subfigure} & 
    \begin{subfigure}{0.14\textwidth}
        \includegraphics[width=\textwidth]{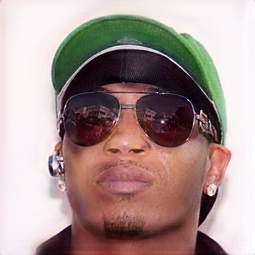}
    \end{subfigure} \\
    
    \rotatebox{90}{\hskip 0.5em \textbf{DiffProtect-fast}} & 
    \begin{subfigure}{0.14\textwidth}
        \includegraphics[width=\textwidth]{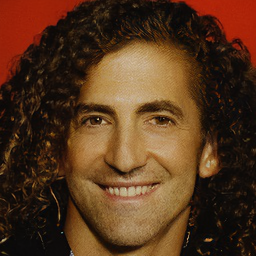}
    \end{subfigure} & 
    \begin{subfigure}{0.14\textwidth}
        \includegraphics[width=\textwidth]{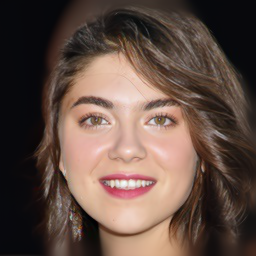}
    \end{subfigure} & 
    \begin{subfigure}{0.14\textwidth}
        \includegraphics[width=\textwidth]{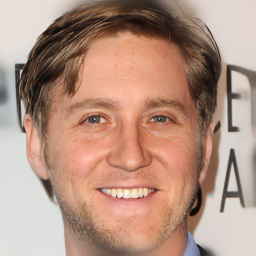}
    \end{subfigure} & 
    \begin{subfigure}{0.14\textwidth}
        \includegraphics[width=\textwidth]{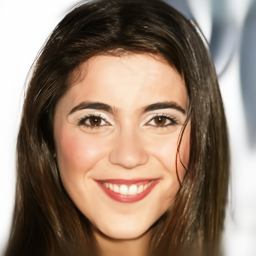}
    \end{subfigure} & 
    \begin{subfigure}{0.14\textwidth}
        \includegraphics[width=\textwidth]{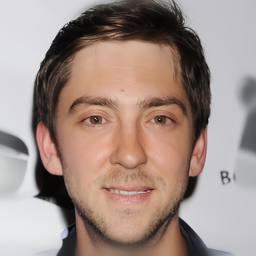}
    \end{subfigure} & 
    \begin{subfigure}{0.14\textwidth}
        \includegraphics[width=\textwidth]{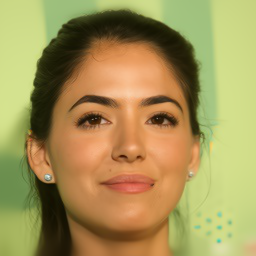}
    \end{subfigure} & 
    \begin{subfigure}{0.14\textwidth}
        \includegraphics[width=\textwidth]{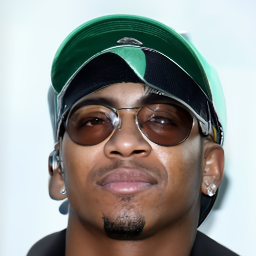}
    \end{subfigure} \\

    \rotatebox{90}{\hskip 1.5em \textbf{DiffProtect}} & 
    \begin{subfigure}{0.14\textwidth}
        \includegraphics[width=\textwidth]{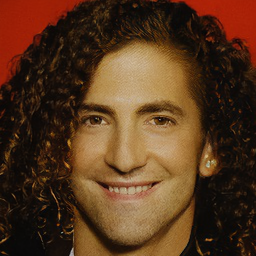}
    \end{subfigure} & 
    \begin{subfigure}{0.14\textwidth}
        \includegraphics[width=\textwidth]{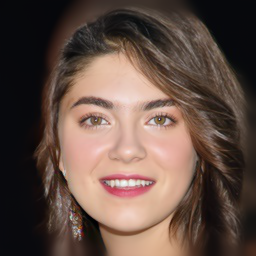}
    \end{subfigure} & 
    \begin{subfigure}{0.14\textwidth}
        \includegraphics[width=\textwidth]{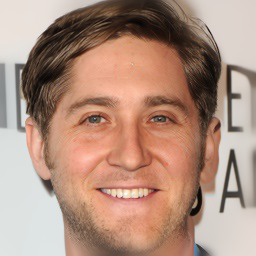}
    \end{subfigure} & 
    \begin{subfigure}{0.14\textwidth}
        \includegraphics[width=\textwidth]{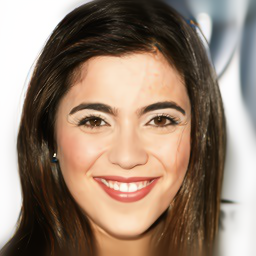}
    \end{subfigure} & 
    \begin{subfigure}{0.14\textwidth}
        \includegraphics[width=\textwidth]{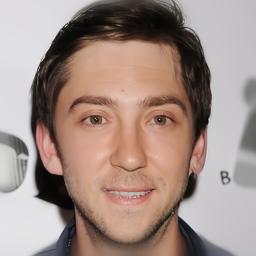}
    \end{subfigure} & 
    \begin{subfigure}{0.14\textwidth}
        \includegraphics[width=\textwidth]{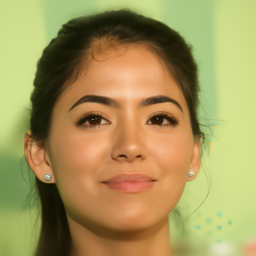}
    \end{subfigure} & 
    \begin{subfigure}{0.14\textwidth}
        \includegraphics[width=\textwidth]{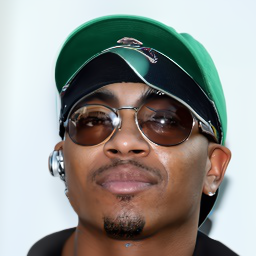}
    \end{subfigure} \\
\end{tabular}}
\caption{Visualizations of the protected face images generated by different face encryption methods on CelebA-HQ.}
\label{fig:supp}
\end{figure*}

\begin{figure*}[t]
\captionsetup[subfigure]{labelformat=empty}
    \centering
    \small
    \setlength{\tabcolsep}{0.5mm}
\scalebox{0.98}{
\begin{tabular}[b]{cccccccc}
    \rotatebox{90}{\hskip 2.5em Input} & 
    \begin{subfigure}{0.14\textwidth}
        \includegraphics[width=\textwidth]{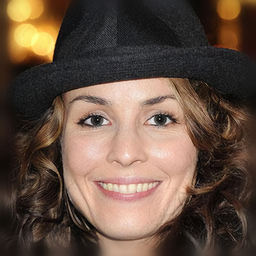}
    \end{subfigure} & 
    \begin{subfigure}{0.14\textwidth}
        \includegraphics[width=\textwidth]{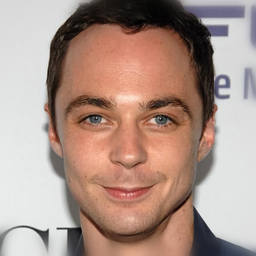}
    \end{subfigure} & 
    \begin{subfigure}{0.14\textwidth}
        \includegraphics[width=\textwidth]{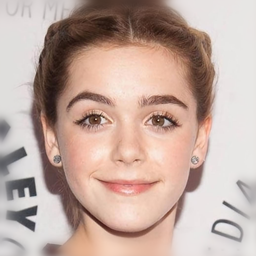}
    \end{subfigure} & 
    \begin{subfigure}{0.14\textwidth}
        \includegraphics[width=\textwidth]{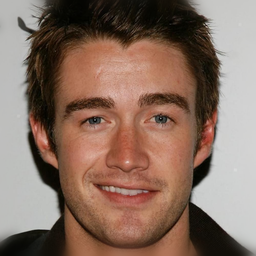}
    \end{subfigure} & 
    \begin{subfigure}{0.14\textwidth}
        \includegraphics[width=\textwidth]{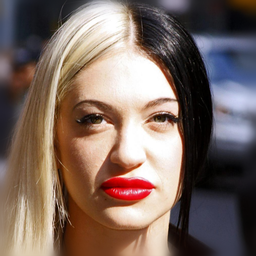}
    \end{subfigure} & 
    \begin{subfigure}{0.14\textwidth}
        \includegraphics[width=\textwidth]{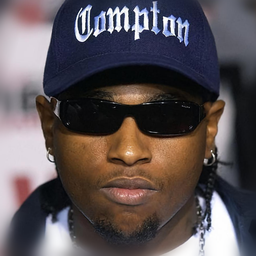}
    \end{subfigure} & 
    \begin{subfigure}{0.14\textwidth}
        \includegraphics[width=\textwidth]{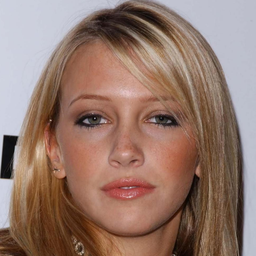}
    \end{subfigure} \\

    \rotatebox{90}{\hskip 1em TIP-IM~\cite{yang2021towards}} & 
    \begin{subfigure}{0.14\textwidth}
        \includegraphics[width=\textwidth]{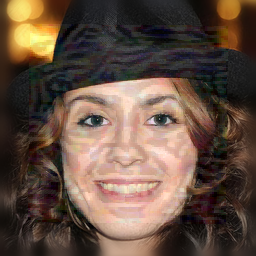}
    \end{subfigure} & 
    \begin{subfigure}{0.14\textwidth}
        \includegraphics[width=\textwidth]{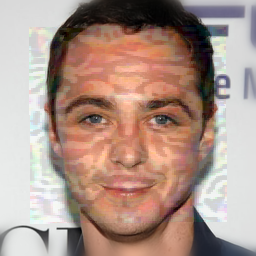}
    \end{subfigure} & 
    \begin{subfigure}{0.14\textwidth}
        \includegraphics[width=\textwidth]{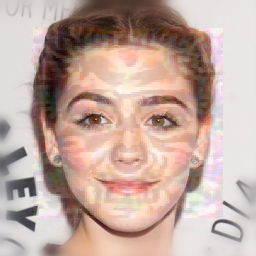}
    \end{subfigure} & 
    \begin{subfigure}{0.14\textwidth}
        \includegraphics[width=\textwidth]{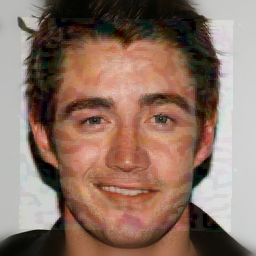}
    \end{subfigure} & 
    \begin{subfigure}{0.14\textwidth}
        \includegraphics[width=\textwidth]{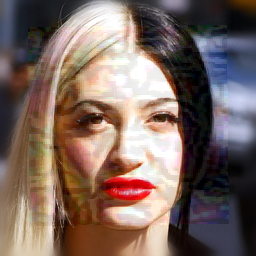}
    \end{subfigure} & 
    \begin{subfigure}{0.14\textwidth}
        \includegraphics[width=\textwidth]{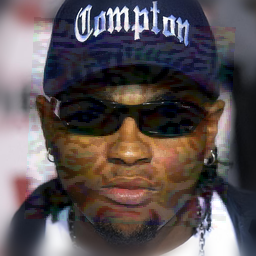}
    \end{subfigure} & 
    \begin{subfigure}{0.14\textwidth}
        \includegraphics[width=\textwidth]{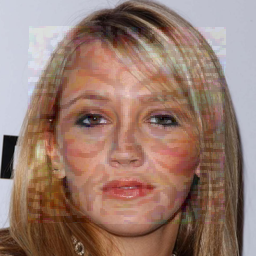}
    \end{subfigure} \\

    \rotatebox{90}{\hskip 0.5em AMT-GAN~\cite{hu2022protecting}} & 
    \begin{subfigure}{0.14\textwidth}
        \includegraphics[width=\textwidth]{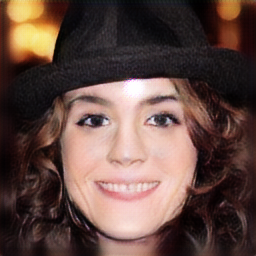}
    \end{subfigure} & 
    \begin{subfigure}{0.14\textwidth}
        \includegraphics[width=\textwidth]{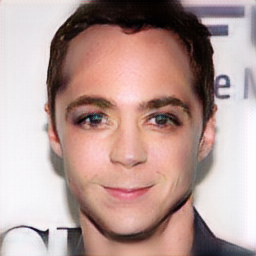}
    \end{subfigure} & 
    \begin{subfigure}{0.14\textwidth}
        \includegraphics[width=\textwidth]{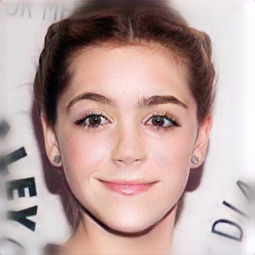}
    \end{subfigure} & 
    \begin{subfigure}{0.14\textwidth}
        \includegraphics[width=\textwidth]{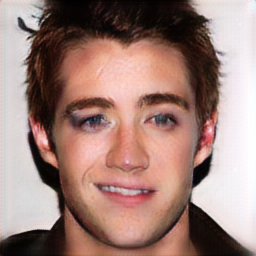}
    \end{subfigure} & 
    \begin{subfigure}{0.14\textwidth}
        \includegraphics[width=\textwidth]{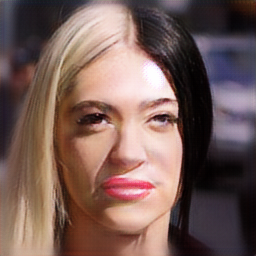}
    \end{subfigure} & 
    \begin{subfigure}{0.14\textwidth}
        \includegraphics[width=\textwidth]{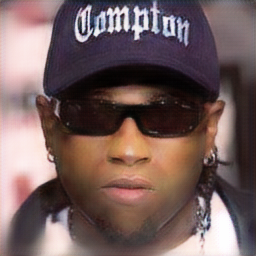}
    \end{subfigure} & 
    \begin{subfigure}{0.14\textwidth}
        \includegraphics[width=\textwidth]{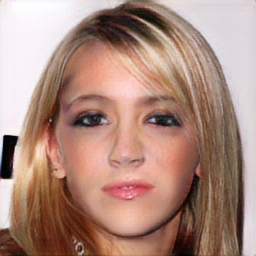}
    \end{subfigure} \\
    
    \rotatebox{90}{\hskip 0.5em \textbf{DiffProtect-fast}} & 
    \begin{subfigure}{0.14\textwidth}
        \includegraphics[width=\textwidth]{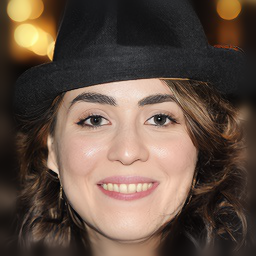}
    \end{subfigure} & 
    \begin{subfigure}{0.14\textwidth}
        \includegraphics[width=\textwidth]{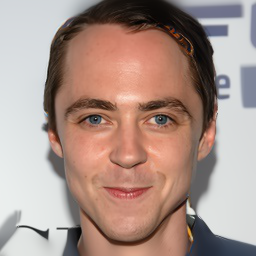}
    \end{subfigure} & 
    \begin{subfigure}{0.14\textwidth}
        \includegraphics[width=\textwidth]{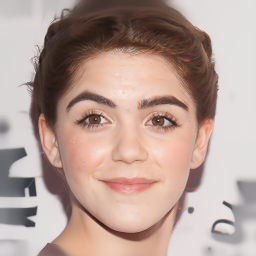}
    \end{subfigure} & 
    \begin{subfigure}{0.14\textwidth}
        \includegraphics[width=\textwidth]{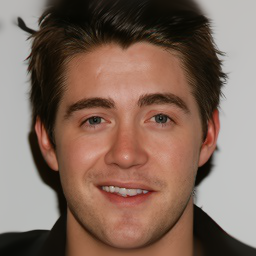}
    \end{subfigure} & 
    \begin{subfigure}{0.14\textwidth}
        \includegraphics[width=\textwidth]{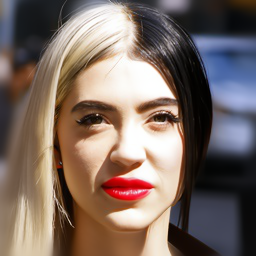}
    \end{subfigure} & 
    \begin{subfigure}{0.14\textwidth}
        \includegraphics[width=\textwidth]{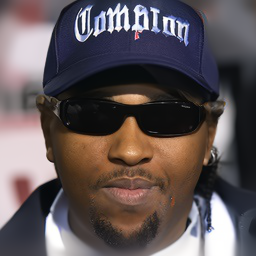}
    \end{subfigure} & 
    \begin{subfigure}{0.14\textwidth}
        \includegraphics[width=\textwidth]{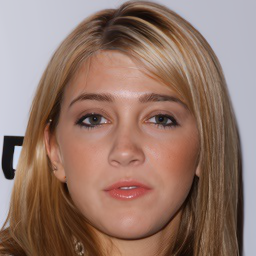}
    \end{subfigure} \\

    \rotatebox{90}{\hskip 1.5em \textbf{DiffProtect}} & 
    \begin{subfigure}{0.14\textwidth}
        \includegraphics[width=\textwidth]{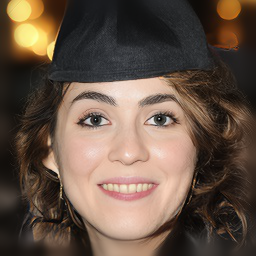}
    \end{subfigure} & 
    \begin{subfigure}{0.14\textwidth}
        \includegraphics[width=\textwidth]{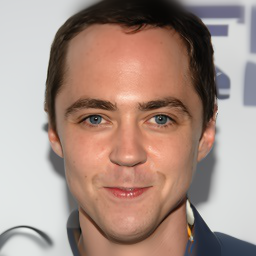}
    \end{subfigure} & 
    \begin{subfigure}{0.14\textwidth}
        \includegraphics[width=\textwidth]{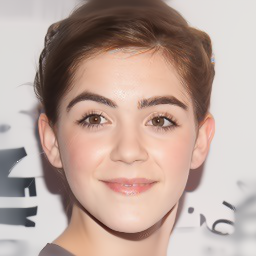}
    \end{subfigure} & 
    \begin{subfigure}{0.14\textwidth}
        \includegraphics[width=\textwidth]{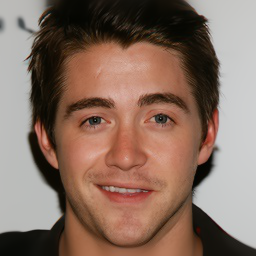}
    \end{subfigure} & 
    \begin{subfigure}{0.14\textwidth}
        \includegraphics[width=\textwidth]{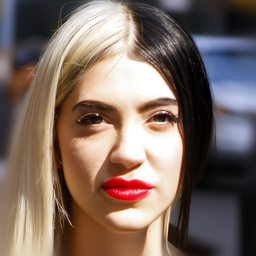}
    \end{subfigure} & 
    \begin{subfigure}{0.14\textwidth}
        \includegraphics[width=\textwidth]{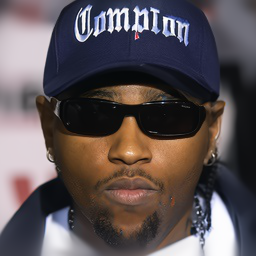}
    \end{subfigure} & 
    \begin{subfigure}{0.14\textwidth}
        \includegraphics[width=\textwidth]{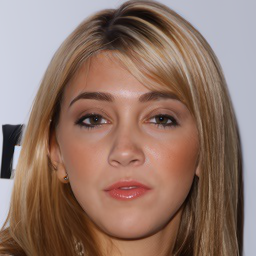}
    \end{subfigure} \\
\end{tabular}}
\caption{Visualizations of the protected face images generated by different face encryption methods on CelebA-HQ.}
\label{fig:supp_2}
\end{figure*}

We provide more examples of protected face images in~\cref{fig:supp,fig:supp_2}. We can observe that DiffProtect produces good-looking protected images with natural and inconspicuous changes to the input images, such as slight changes in facial expressions. It works well across genders, ages, and races, and in some cases even makes the images look more attractive. Compared to TIP-IM, the protected face images generated by DiffProtect have no obvious noise pattern as we only perturb the semantic codes and generate the images through a conditional DDIM. Compared to AMT-GAN, DiffProtect can better preserve image styles and details and does not require training a target-specific model for each identity. 


\end{document}